%% file: main.tex
\documentclass[twocol]{ametsocV6.1}

\usepackage{graphicx}
\usepackage{blindtext}
\usepackage{float}
\usepackage{booktabs}
\usepackage[utf8]{inputenc} % enable direct typing of ä, ö, ü
\usepackage[T1]{fontenc} % same here

\usepackage{bbm} %indicator
\usepackage{multirow} % fancy tables

\makeatletter
\newcommand\thefontsize[1]{{#1 The current font size is: \f@size pt\par}}
\makeatother

\title{
Postprocessing of Ensemble Weather Forecasts \\Using Permutation-invariant Neural Networks
}

\authors{
Kevin Höhlein,\aff{a}\correspondingauthor{Kevin Höhlein, kevin.hoehlein@tum.de}% 
Benedikt Schulz,\aff{b}%
Rüdiger Westermann,\aff{a}%
and Sebastian Lerch\aff{b,c}
}

\affiliation{%
\aff{a}{Technical University of Munich}\\
\aff{b}{Karlsruhe Institute of Technology}\\
\aff{c}{Heidelberg Institute for Theoretical Studies}
}
 
\input{abstract}

\begin{document}

\maketitle

\input{introduction}
\input{related_work}
\input{methods}
\input{dataset}

\input{results}
\input{discussion}

% \clearpage

\acknowledgments
\input{acknowledgments}

\datastatement
The case study on surface temperature postprocessing is based on the EUPPBench dataset, which is publicly available. See \citet{demaeyer2023euppbench} for details. The wind gust dataset is proprietary but can be obtained from the DWD for research purposes. Code with implementations of all methods is publicly available \citep{hoehlein2023code}.

\input{appendix}

\clearpage

\bibliographystyle{ametsocV6}
\bibliography{references}

\end{document}

%% file: abstract.tex
\abstract{
Statistical postprocessing is used to translate ensembles of raw numerical weather forecasts into reliable probabilistic forecast distributions. 
In this study, we examine the use of permutation-invariant neural networks for this task. In contrast to previous approaches, which often operate on ensemble summary statistics and dismiss details of the ensemble distribution, we propose networks that treat forecast ensembles as a set of unordered member forecasts and learn link functions that are by design invariant to permutations of the member ordering. We evaluate the quality of the obtained forecast distributions in terms of calibration and sharpness and compare the models against classical and neural network-based benchmark methods. In case studies addressing the postprocessing of surface temperature and wind gust forecasts, we demonstrate state-of-the-art prediction quality. 
To deepen the understanding of the learned inference process, we further propose a permutation-based importance analysis for ensemble-valued predictors, which highlights specific aspects of the ensemble forecast that are considered important by the trained postprocessing models. 
Our results suggest that most of the relevant information is contained in a few ensemble-internal degrees of freedom, which may impact the design of future ensemble forecasting and postprocessing systems.
}

%% file: introduction.tex
\section{Introduction}

Operational weather forecasting relies on numerical weather prediction (NWP) models. Since such models are subject to multiple sources of uncertainty, such as uncertainty in the initial conditions or model parameterizations, a quantification of the forecast uncertainty is indispensable.
To achieve this, NWP models generate a set of deterministic forecasts, so-called ensemble forecasts, based on different initial conditions and variations of the underlying physical models.
Since these forecasts are subject to systematic errors such as biases and dispersion errors, statistical postprocessing is used to enhance their reliability \citep[see, e.g.,][]{Vannitsem2018}.
Recently, machine learning (ML) approaches for statistical postprocessing have shown superior performance over classical methods. For instance, \citet{Rasp2018} propose a distribution regression network~(DRN) which predicts the parameters of a temperature forecast distribution from a suitable family of parametric distributions. In subsequent work, \citet{Schulz2022} found that shallow multi-layer perceptrons~(MLPs) with forecast distributions of different flexibility achieve state-of-the-art results in postprocessing wind gust ensemble forecasts. 

An ensemble forecast consists of multiple separate member forecasts, which are generated by repeatedly running NWP simulations with different model parameterizations and initial conditions. Typically, the configurations of different runs are sampled randomly from an underlying distribution of plausible simulation conditions, obtained, e.g., from uncertainty-aware data assimilation. The member forecasts can then be seen as identically distributed and interchangeable random samples from a distribution of possible future weather states. In this setting, statistical postprocessing of ensemble forecasts can be phrased as a prediction task on unordered predictor vectors and requires solutions that are tailored to match the predictor format. Specifically, member interchangeability demands that the predictions of a well-designed postprocessing system should not be affected by permutations, i.e.\ shuffling, of the ensemble members. Systems that satisfy this requirement are called permutation invariant. 
Established postprocessing methods rely on basic summary statistics of the raw ensemble forecast to inform the estimation of the postprocessed distribution and are thus permutation invariant by design. However, especially in large ensembles, the details of the distribution may carry valuable information for postprocessing, and a more elaborate treatment of the inner structure of the raw forecast ensembles may help to improve forecast accuracy for example in ambiguous forecast situations, where summary-based methods fail to evaluate the likelihood of different weather patterns accurately.

While studies have started to explore how specialized model architectures can help to improve postprocessing only recently~\citep{Bremnes2020,mlakar2023ensemble,ben2023improving}, ML provides a variety of further approaches to enforcing permutation invariance in data-driven learning. Motivated by the success of permutation-invariant neural network~(NN) architectures in representation learning, anomaly detection or set classification~\citep[e.g.,][]{Ravanbakhsh2016,Zaheer2017,Lee2019,Sannai2019,zhang2019fspool}, where the models profit from the ability to extract concise feature representations from unordered data, permutation-invariant NNs appear as promising candidates for improving ensemble postprocessing.

\subsection*{Contribution}

In this study, we investigate the capabilities of different permutation-invariant NN architectures for univariate postprocessing of station predictions. We evaluate the proposed models on two exemplary station-wise postprocessing tasks with different characteristics. The ensemble-based network models are compared to classical methods and basic NNs which operate only on ensemble summary statistics but are trained under identical predictor conditions otherwise. 
We further assess how much of the predictive information is carried within the details of the ensemble distribution, and how much of the model skill arises from other factors. To shed light on the sources of model skill, we propose an ensemble-oriented feature importance analysis and study the effect of ensemble-internal degrees of freedom using conditional feature permutation. 

%% file: related_work.tex
\section{Related work\label{sec:relwork}}

\subsection{Statistical postprocessing of ensemble forecasts\label{sec:relwork:pp}}

One of the most popular methods for statistical postprocessing of ensemble forecasts is ensemble model output statistics \citep[EMOS;][]{Gneiting2005}, which performs a distributional regression based on a suitable family of parametric distributions and summary statistics of the ensemble. 
Due to its simplicity, EMOS has been applied to a wide range of weather variables including temperature \citep{Gneiting2005}, wind gusts \citep{Pantillon2018}, precipitation \citep{Scheuerer2014} and solar radiation \citep{Schulz2021}. Following the simple statistical EMOS approach, the success of ML methods \citep{Taillardat2016, Messner2017}, which are able to incorporate additional information and learn more complex patterns, have motivated the use of modern NN-methods.
First NN-based approaches are DRN \citep{Rasp2018} as an extension of the EMOS framework, and the Bernstein quantile network \citep[BQNs;][]{Bremnes2020} that provides a more flexible forecast distribution. In \citet{Schulz2022}, NN-based approaches were adapted towards the prediction of wind gusts and outperformed classical methods. Recently, research has shifted towards the use of more sophisticated network architectures. Examples include convolutional NNs that incorporate spatial NWP output fields \citep{Scheuerer2020,gronquist2021deep,Veldkamp2021,horat_lerch_2023},
and generative models to produce multivariate forecast distributions \citep{Hemri2021,Chen2022}. 

Only recently, \citet{mlakar2023ensemble} have proposed NN models that explicitly admit the use of ensemble-structured predictors by employing a dynamic attention mechanism. The resulting models perform best in the benchmark study of \citet{demaeyer2023euppbench}. 
\citet{mlakar2023ensemble} address postprocessing with similar methods as this work, but do not focus explicitly on comparing different network design patterns for inference based on ensemble-valued predictors.
In orthogonal work, \citet{finn2021self} and \citet{ben2023improving} apply transformer-based NNs to ensemble postprocessing. In contrast to this study, both approaches focus on gridded predictor data, thus relying on different network architectures, and postprocess ensembles in a member-by-member fashion, whereas this work concentrates on distributional regression.

For a general review of statistical postprocessing of weather forecasts, we refer to \cite{Vannitsem2018}, a review of recent developments and challenges can be found in \citet{Vannitsem2021} and \citet{Haupt2021}.

\subsection{Neural network architectures for regression on set-structured data}

From an ML perspective, postprocessing of ensemble forecasts can be phrased as a regression task on set-structured predictors.
Multiple studies have demonstrated that dedicated permutation-invariant NN architectures can help to improve prediction quality and generalization in diverse learning problems \citep[e.g.,][]{vinyals2015order,Lyle2020}, thus motivating the exploration of permutation-invariant NNs also for ensemble postprocessing. Early works on permutation-invariant layers for NNs \citep{Ravanbakhsh2016} and pooling-based permutation-invariant NNs \citep{edwards2016towards} were followed by the more comprehensive framework \emph{DeepSets} \citep{Zaheer2017}, which encompasses some of the most common design patterns for ML inference on set-structured predictors. Due to its generality, \emph{DeepSets} is selected as one of the representative learning approaches in this study and is further discussed in Section~\ref{sec:networks}\ref{sec:deepsets}. \citet{soelch2019aggregations} highlight that architectural improvements, such as the use of more expressive pooling functions, may enhance model performance, which we consider in the design of the model architectures for postprocessing. 

An alternative approach to permutation-invariant inference has been proposed by \cite{Lee2019}, who use (multi-head) attention functions~\citep{Vaswani2017} for permutation-invariant inference on set-valued data. Attention-based models, also known as transformers, have proven powerful in a variety of computer vision tasks~\citep[e.g.,][]{khan2022transformers} as well as postprocessing~\citep[][see Section~\ref{sec:relwork}\ref{sec:relwork:pp}]{finn2021self,ben2023improving} and are thus considered as a second paradigm for building permutation-invariant NNs. 

\subsection{Machine learning explainability and feature importance}
 
ML explainability has attracted substantial interest throughout the last decade \citep[for recenct surveys see, e.g.,][]{guidotti2018survey, linardatos2020explainable, burkart2021survey} and is increasingly adopted in the earth-system sciences~\citep[e.g.,][]{reichstein2019deep, hohlein2020comparative, labe2021detecting, farokhmanesh2023deep} to gain understanding on the reasoning mechanisms behind ML-based inference approaches. The most relevant approaches for this work are based on permutation feature importance~\citep[PFI;][]{breiman2001random}, which aims to assess the (relative) importance of different predictors for inference. In PFI, relevance scores are assigned to the predictors based on the accuracy loss after permuting the predictor values within the test dataset, and have been applied in the postprocessing before~\citep[e.g.][]{Rasp2018,Schulz2022} with a focus on scalar-valued predictors. In this work, we propose a conditional PFI measure for ensemble-valued predictors, which allows attributing importance values to different aspects of the ensemble-internal variability. Conditional perturbation measures have been considered in earlier works \citep[e.g.,][]{strobl2008conditional,molnar2023model}, where the conditioning is used to evaluate the importance of specific predictors in the context of the remaining predictors. By contrast, our approach addresses specifically the distribution characteristics of the ensemble-valued predictors encountered in postprocessing.

%% file: methods.tex
\section{Benchmark methods and forecast distributions\label{sec:benchmark}}

\subsection{Assessing predictive performance}

We evaluate probabilistic forecasts based on the paradigm of \cite{Gneiting2007probabilistic}, i.e., a forecast should maximize sharpness subject to calibration. Both sharpness and calibration can be assessed quantitatively using proper scoring rules \citep{Gneiting2007scoring}. A popular choice is the continuous ranked probability score \citep[CRPS;][]{Matheson1976} %which is given by
\begin{eqnarray*}
	\text{CRPS} (F, y) = \int_{-\infty}^{\infty} \left( F(z) - \mathbbm{1} \lbrace y \leq z \rbrace \right)^2 \text{d}z,
\end{eqnarray*}
wherein $y \in \mathbb{R}$ is the observed value, $F$ the cumulative distribution function (CDF) of the forecast distribution, and $\mathbbm{1}$ the indicator function.
The CRPS can be computed analytically for a wide range of distributions including the truncated logistic distribution and probabilistic forecasts in ensemble form \citep{Jordan2019}. 

In addition to the CRPS, we assess calibration based on the empirical coverage of prediction intervals (PIs) derived from the forecast distribution, and sharpness on the corresponding length. Under the assumption of calibration, the observed coverage of a PI should match the nominal level, and a forecast is sharper the smaller the length of the PI. 
In line with \citet{Schulz2022}, we choose the PI level based on the size of the underlying ensemble. For an ensemble of size $M\in\mathbb{N}$, this gives rise to a PI with nominal level $(M-1)/(M+1)$.

Further, we qualitatively assess calibration based on (unified) probability integral transform (PIT) histograms \citep{Gneiting2014,vogel2018skill}. While a flat histogram indicates that the forecasts are calibrated, systematic deviations indicate miscalibration.
For more details on the evaluation of probabilistic forecasts, we refer to \citet{Gneiting2014}.

\subsection{Distributional regression with parametric forecast distributions (EMOS, DRN)}

In this study, we consider postprocessing of the ensemble forecast for a real-valued random variable $Y$ as a distributional regression task on ensemble-structured predictors. We focus on the case of station-wise forecasts, which are given as prediction vectors $\bm{x}\in \mathbb{R}^p$, each comprising the predictions of $p$ scalar-valued meteorological variables, such as surface temperature or 10-m wind speed at a station site. Typically, one of the forecast variables corresponds directly to the target variable $Y$ and is thus termed the primary prediction. An $M$-member ensemble forecast $X = \left\{\bm{x}_1, ..., \bm{x}_M\right\} \subset \mathbb{R}^p$ is composed of $M$ such prediction vectors, which respresent samples from the predicted distribution of future weather states. The space of $M$-member ensemble forecasts will be denoted as $\left[\mathbb{R}^p\right]_M$.

Within the (parametric) distributional regression framework, the parameter vector $\bm{\theta}$
of a parametric distribution $\mathcal{F}_\theta$ is linked to the predictors via a function that is estimated by minimizing a proper scoring rule. The underlying model can be written as
\begin{eqnarray}
	Y \mid X \sim \mathcal{F}_{\bm{\theta}}, \quad \bm{\theta} = g\!\left( X \right) \in \Theta, \label{eq:emos_distribution}
\end{eqnarray}
where $g: \left[\mathbb{R}^p\right]_M \rightarrow \Theta$ is called the link function and  $\Theta \subseteq \mathbb{R}^{D}$ denotes the $D$-dimensional parameter space corresponding to $\mathcal{F}_\theta$. 

For EMOS, $g$ is defined as a generalized affine-linear function of ensemble summary statistics, such as ensemble mean or standard deviation, and provides only limited flexibility for distribution estimation. DRN \citep{Rasp2018,Schulz2022}, in contrast, admits the data-driven estimation of arbitrary link functions using NNs, thus increasing the learning ability. The forecast distribution as well as the underlying proper scoring rule used for optimization are two implementation choices. 

\subsection{Flexible distribution estimator (BQN)}

Distributional regression methods based on a parametric forecast distribution are robust but lack flexibility as they are bound to the parametric distribution family of choice. Typical choices of forecast distributions include the normal \citep{Gneiting2005,Rasp2018}, logistic \citep{Schulz2022} or generalized extreme value distribution \citep{lerch_thorarinsdottir_2013,Scheuerer2014}. They all lack the ability to express multi-modalities that are required, e.g., when different weather patterns occur. Hence, methods that do not rely on parametric assumptions have been proposed in the postprocessing literature. Examples are the direct adjustment of the ensemble members \citep{VanSchaeybroeck2015} or quantile regression forests \citep{Taillardat2016}. 
BQN \citep{Bremnes2020} models the forecast distribution as a quantile function, which is represented as a linear combination of Bernstein (basis-)polynomials of degree $d\in\mathbb{N}$ with variable mixing coefficients $\bm{\alpha} = (\alpha_0, ..., \alpha_d) \in \mathbb{R}^{d+1}$, such that $\alpha_0 \leq ... \leq \alpha_d$.
The inference network is designed to output parameters $\bm{\theta}$ that parameterize the mixing coefficients, i.e., $\bm{\alpha} = \bm{\alpha}\!\left(\bm{\theta}\right)$. 
In contrast to DRN, this formulation offers increased flexibility for modeling multi-modality, while requiring hard upper and lower bounds for the values of the forecast variable. For BQN models, the optimization is guided by an average of quantile scores \citep{Koenker1978}, which can be seen as a discrete approximation of the CRPS \citep{Gneiting2011}.
For the evaluation of BQN forecasts, we generate an ensemble of equidistant quantiles analogous to \citet{Schulz2022}.
In the original implementation, the link function of BQN is specified as an NN, which receives a sorted sequence of univariate ensemble predictors as its input~\citep{Bremnes2020}. \cite{Schulz2022} augment this approach by using ensemble-valued predictors for the predictor of the target variable and ensemble means for additional auxiliary predictor variables. Despite admitting permutation-invariant inference, both model variants are constrained to processing ensembles of fixed size. To alleviate this limitation, we avoid the sorting operation in this work and inform BQN models analogous to DRN using ensemble summary statistics. A comparison of both variants is conducted in the supplementary materials, demonstrating the equivalence of the approaches.

\subsection{Use of auxiliary predictors\label{sec:auxpred}}

In addition to the predictions of the target variable, most algorithms use auxiliary information to improve the prediction performance (see Table~\ref{tab:predictors}). 
We distinguish between ensemble-valued and scalar-valued predictors, where ensemble-valued predictors vary between different members and scalar-valued predictors do not. In the ensemble-valued case, we differentiate the primary prediction from auxiliary predictions of other meteorological variables.
For either of these, postprocessing models can have access to the full set of ensemble values or only to summary statistics.

Scalar-valued predictors refer to contextual information, such as station-specific coordinates and orography details (cf.\ Table~\ref{tab:predictors}: station pred.), as well as to temporal information, such as the day of the year. We consider only models that are trained on predictions for specific initialization and lead times, such that information about the diurnal cycle is not required. While most approaches include the scalar predictors explicitly as features in the regression process, EMOS takes advantage of categorical location and time information implicitly by fitting independent models for each station and month~\citep{Schulz2022}. BQN- and DRN-type models are trained separately for each lead time, but employ a learned station embedding \citep{Rasp2018,Schulz2022} to share the same model between different station sites.
Notably, the permutation-invariant models (cf.\ Table~\ref{tab:predictors}: perm.-inv.) considered in this study have access to the richest predictor pool. A complete list of model inputs on the parameter level can be found in Tables~\ref{tab:predictors:gusts}, \ref{tab:predictors:eupp} and \ref{tab:predictors:joint} in Appendix~A. 

\begin{table*}[]
    \centering
    \caption{Predictor utilization by postprocessing methods. Methods used in this study are indicated by \emph{ours}.}
    \begin{tabular}{l c c c c }
        \toprule
        Predictors &  \multicolumn{2}{c}{Ensemble-valued} & \multicolumn{2}{c}{Scalar-valued}  \\
        \midrule
        Method & Primary prediction & Auxiliary predictions & Spatial & Temporal \\
        \midrule 
        EMOS \citep[][ours]{Schulz2022} & mean + std. dev. & -- & \multicolumn{2}{c}{different models per station and month} \\
        \midrule
        BQN \citep{Bremnes2020} & ensemble (sorted) & -- & station embed. & -- \\
        BQN \citep{Schulz2022} & ensemble (sorted) & mean & station pred. + embed.  & day of year \\
        BQN (ours) & mean + std. dev. & mean & station pred. + embed.  & day of year \\
        DRN \citep[][ours]{Schulz2022} & mean + std. dev. & mean & station pred. + embed.  & day of year \\
        \midrule
        Perm.-inv. DRN + BQN (ours) & ensemble (perm.-inv.) & ensemble (perm.-inv.) & station pred. + embed. & day of year \\
        \bottomrule
    \end{tabular}
    \label{tab:predictors}
\end{table*}

\section{Permutation-invariant neural network architectures\label{sec:networks}}

\begin{figure}
    \centering
    \includegraphics[width=.4\textwidth]{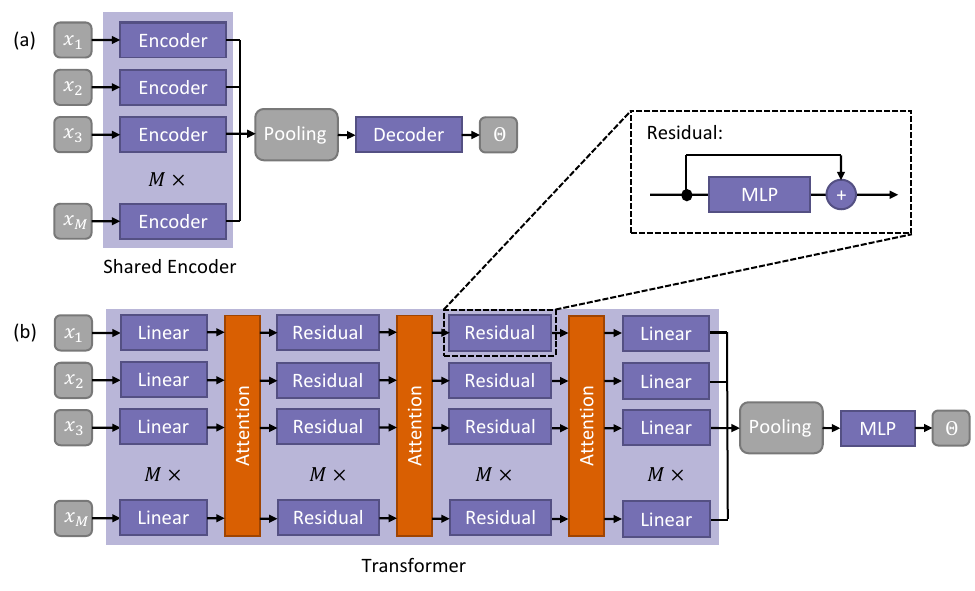}
    \caption{Set pooling architecture (a), consisting of encoder and decoder MLPs, and set transformer (b), featuring attention blocks and intermediate MLPs with residual connections. While the encoder-decoder architecture admits interactions between members only inside the pooling step, the set transformer admits information transfer between the members in each attention step.}
    \label{fig:architectures}
\end{figure}

From the variety of permutation-invariant model architectures, we select two representative approaches, \emph{set pooling architectures} and \emph{set transformers}, which we adapt for distributional regression. Compared with the benchmark methods of Section~\ref{sec:benchmark}, the proposed networks replace the summary-based ensemble processing while the parameterization of the forecast distributions remains unchanged.
A schematic comparison of both permutation-invariant architectures is shown in Fig.~\ref{fig:architectures}. 

\subsection{Set pooling architectures\label{sec:deepsets}}

Set pooling architectures~\citep{Zaheer2017}, also known as \emph{DeepSets}, achieve permutation invariance via extraction and permutation-invariant summarization of learned latent features. The features are obtained by applying an encoder MLP to all ensemble members separately, followed by a permutation-invariant pooling function and a decoder MLP, which outputs the distribution parameters $\theta$. 
Due to the division into encoding, pooling, and decoding, 
we will thus use the names \emph{set pooling} and \emph{encoder-decoder} (ED) architecture synonymously. 

In experiments, we considered different variants of ensemble summarization based on average and extremum pooling, as well as adaptive pooling functions based on an attention mechanism \citep{Lee2019,soelch2019aggregations},
discussed below. Overall, we find that the pooling mechanism is of minor importance. Detailed comparisons are thus deferred to the supplementary materials and all subsequent experiments use attention-based pooling consistently.

\subsection{Set Transformer\label{sec:transformer}}

Set transformers~\citep{Lee2019} are NNs, that model interactions between set members via self-attention. \emph{Attention} is a form of nonlinear activation function, in which the relevance of the inputs is determined via a matching of input-specific key and query vectors. \emph{Multi-head attention} allows the model to attend to multiple key patterns in parallel \citep{Vaswani2017}. \citet{Lee2019} combine multi-head attention with a member-wise NN to build a permutation-invariant set-attention block, from which a set transformer is constructed by stacking multiple instances. Set transformers apply straight-forwardly to ensemble data and 
can exploit all aspects of the available ensemble dataset by allowing for information exchange between ensemble members early in the inference process. We construct a set transformer by using three set-attention blocks with 8 attention heads \citep{Vaswani2017,Lee2019}. Each block comprises a separate MLP with two hidden layers. Additionally, the first set-attention block is preceded by a linear layer to align the channel number of the ensemble input with the hidden dimension of the set-attention blocks. To construct vector-valued predictions from set-valued inputs, \citet{Lee2019} propose attention-based pooling, in which the output query vectors are implemented as learnable parameters. 
After pooling, the final prediction $\theta$ is obtained by applying another two-layer MLP.

%% file: dataset.tex
\section{Data} \label{sec:data}

We evaluate the performance of the proposed models in two postprocessing tasks using the datasets described in Table \ref{tab:data}. An overview of the predictor and target variables is provided in Appendix~A. 

\begin{table*}[t]
	\caption{Overview of the data used in the postprocessing applications described in Section~\ref{sec:data}.}
	\label{tab:data}
	\centering
	% \scalebox{0.62}{ 
 \input{table_data.txt}
 % }
\end{table*}

\subsection{Wind gust prediction in Germany}

In the first case study, we employ our methods for station-wise postprocessing of wind gust forecasts using a dataset that has previously been used in \cite{Pantillon2018} and \cite{Schulz2022}. The ensemble forecasts are based on the COSMO ensemble prediction system \citep[COSMO-DE;][]{Baldauf2011} and consist of 20 members forecasts, simulated with a horizontal resolution of 2.8\,km. The forecasts are initialized at 00 UTC, and we consider the lead times 6, 12, and 18h.
Other than wind gusts, the dataset comprises ensemble forecasts of several meteorological variables, such as temperature, pressure, precipitation, and radiation. An overview of all predictors is shown in Table~\ref{tab:predictors:gusts} (Appendix A). The predictions are verified against observations measured at 175 stations of the German weather service (Deutscher Wetterdienst; DWD). Forecasts for the individual weather stations are obtained from the closest grid point. The time period of the forecast and observation data starts on 9 December 2010 and ends on 31 December 2016. The models use the data from 2010-2015 for model estimation, using 2010-2014 as training and 2015 as validation period. The forecasts are then verified in 2016. As in \cite{Schulz2022}, each lead time is processed separately. 

As detailed in~\cite{Schulz2022}, a minor caveat is caused by a non-trivial substructure of the forecast ensembles. The 20-member ensembles constitute a conglomerate of four sub-ensembles, which are generated with slightly different model configurations. While this formally violates the assumption of statistical interchangeability of the members, the sub-ensembles are sufficiently similar to justify the application of permutation-invariant models. 

For the benchmark methods EMOS and DRN, we use the exact same forecasts as in \cite{Schulz2022}, both estimating the parameters of a truncated logistic distribution by minimizing the CRPS, see their Section 3 for details. BQN is adapted as described in Section~\ref{sec:benchmark} and Table~\ref{tab:predictors}. 

\subsection{Temperature forecasts from the EUPPBench dataset}

In a second example, we postprocess ensemble forecasts of surface temperature using a subset of the EUPPBench postprocessing benchmark dataset \citep{demaeyer2023euppbench}. EUPPBench provides paired forecast and observation data from two sets of samples. The first part consists of 20 years of reforecast data (1997 - 2016) from the Integrated Forecasting System (IFS) of the ECMWF with 11 ensemble members. Mimicking typical operational approaches, the reforecast dataset is used as training data, complemented by additional two years (2017 and 2018) of 51-member forecasts as test data. EUPPBench comprises sample data from multiple European countries -- Austria, Belgium, France, Germany and the Netherlands -- 
which are publicly accessible via the CliMetLab API~\citep{climetlab2022}. Additional data for Switzerland can be requested from the Swiss weather service but is not used in this study. EUPPBench constitutes a comprehensive dataset of samples over a long time period. In contrast to the wind gust forecasts, the EUPPBench ensemble members are exchangeable, so that permutation-invariant model architectures are optimally suited.
%for working with the data. 

Deviating from the EUPPBench convention, models are tested on the 51-member forecasts, and  the last 4 years of the reforecast dataset are considered as an independent test set of 11-member forecast samples. This allows us to assess the generalization capabilities of the ensemble-based postprocessing models on data equivalent to the training data, as well as on data with larger ensemble sizes. Furthermore, we use the full set of available surface- and pressure-level predictor variables, whereas the original EUPPBench task is restricted to using only surface temperature data. While this design choice hinders the direct comparison of the evaluation metrics in this paper with the original EUPPBench models, it enables a more comprehensive assessment of the relative benefits of using summary-based vs.\ ensemble predictors. An overview of the predictors can be found in Table~\ref{tab:predictors:eupp} (Appendix A). From the pool of available forecast lead times, we select 24h, 72h, and 120h for a closer analysis.

Unlike previous postprocessing applications for temperature \citep[e.g.,][]{Gneiting2005,Rasp2018}, we employ a zero-truncated logistic distribution as parametric forecast distribution for DRN, instead of a zero-truncated normal, as preliminary tests showed a slightly superior predictive performance of the logistic distribution pattern (see supplementary material for details). The zero-censoring arises from the use of the Kelvin scale for measuring temperatures and allows the use of the same model configuration for both temperature and wind gust predictions. In particular, the EMOS and DRN benchmark approaches are identical for both data sets.

%% file: table_data.txt
\begin{tabular}{lrr} 
\toprule 
Dataset & Wind gust forecasts & EUPPbench (re)forecasts \\ 
\midrule 
% \multicolumn{5}{l}{\textit{Normalization}} \\ 
Underlying NWP model & COSMO-DE-EPS & ECMWF-IFS  \\ 
Initialization time & 00 UTC & 00 UTC  \\ 
Ensemble size $M$ & 20 & Reforecasts: 11\\
& & Forecasts: 51\\
Predicted ensemble forecast quantities $p$ & 61 & 28 \\ 
Region & Germany & Central Europe \\
Stations & 175 & 117 \\
Lead times considered in h & 6, 12, 18 & 24, 72, 120 \\
Training samples & 315,000 & 374,000 \\
Test samples & 63,000 & Reforecasts: 97,000\\
& & Forecasts: 85,000\\
\bottomrule 
\end{tabular}

%% file: results.tex
\section{Performance evaluation\label{sec:eval}}

For each of the postprocessing methods, we generated a pool of 20 networks in each forecast scenario. To ensure a fair comparison to the benchmark methods, we follow the approach from \cite{Schulz2022preprint,Schulz2022}, who build an ensemble of 10 networks and combine the forecasts via quantile aggregation. Hence, we draw 10 members from the pool and repeat this procedure 50 times to quantify the uncertainty of sampling from the general pool. 
For all model variants and resamples, we select those configurations as the final forecast that yield the lowest CRPS on the validation set. Details on hyperparameter settings are listed in Appendix B  and tuning procedures are discussed in the supplementary materials.
For both datasets, we compute the average CRPS, PI length, and PI coverage for the different forecast lead times based on the respective test datasets. The average is calculated over the resamples of the aggregated network ensembles. In what follows, the prefixes ED and ST refer to pooling-based encoder-decoder models and set transformers, respectively, and suffixes DRN and BQN indicate the parameterization of the forecast distribution. The model categories DRN and BQN without additional prefixes refer to the benchmark models based on summary statistics.

\subsection{Wind gust forecasts}
 
\begin{figure*}
    \centering
    \includegraphics[width=\textwidth]{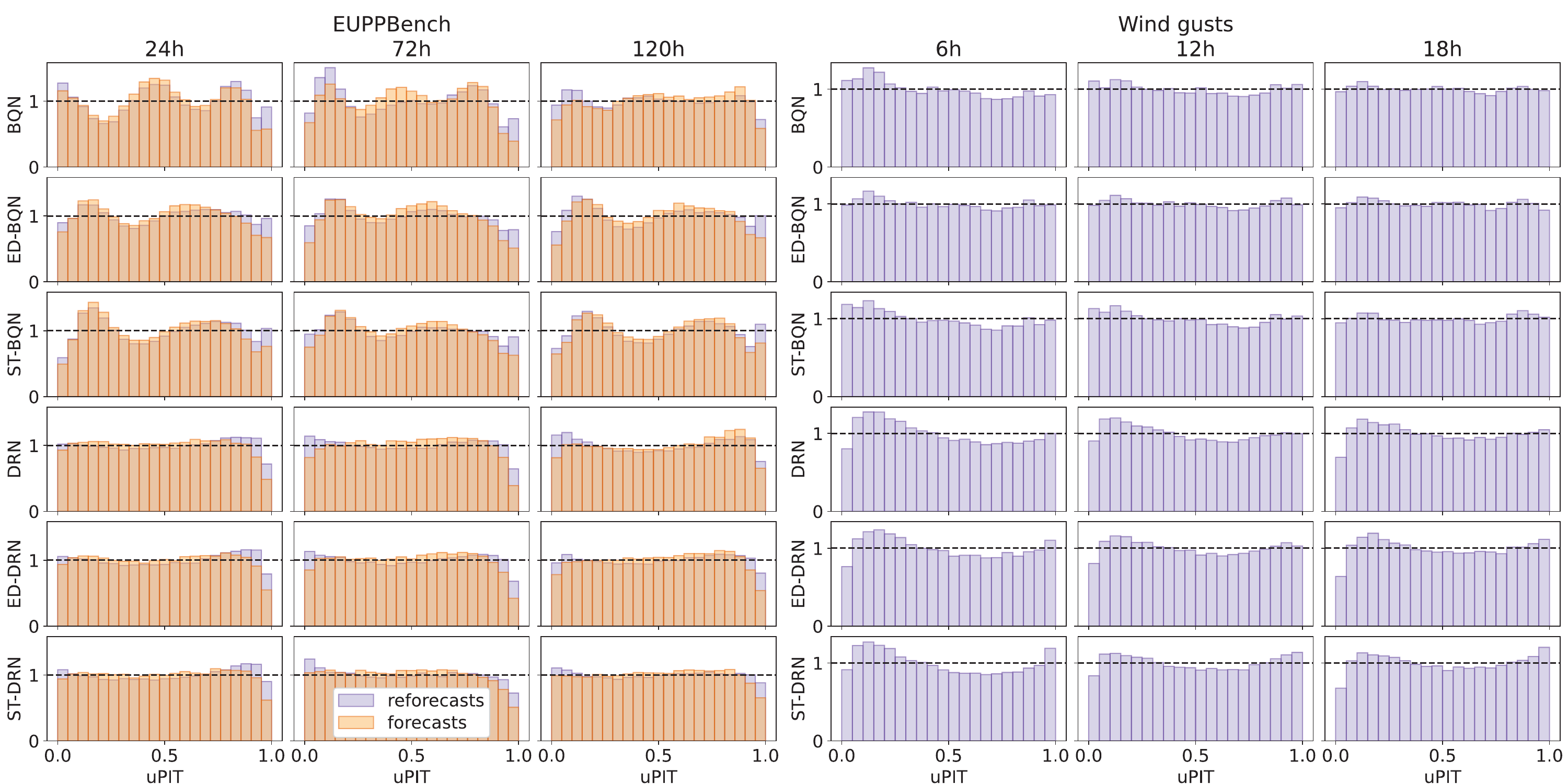}
    \caption{
    PIT histograms of the postprocessing models for EUPPBench 11-member reforecast and 51-member forecast ensembles (left) and 20-member wind gust forecasts (right), 
    }
    \label{fig:eval:calibration}
\end{figure*}

Table~\ref{tab:eval:gusts} shows the quantitative evaluation for lead times 6h, 12h, and 18h. All permutation-invariant model architectures perform similarly to the DRN and BQN benchmarks and outperform both the EPS and conventional postprocessing via EMOS, thus achieving state-of-the-art performance for all lead times. Further, the PI lengths and coverages are similar to those of the benchmark methods with the same forecast distribution, indicating that the ensemble-based models achieve approximately the same level of sharpness as the benchmark networks while being well-calibrated.
Note that the underlying distribution type should be taken into account when comparing the sharpness of different postprocessing models based on the PI length, as the DRN and BQN forecast distributions exhibit different tail behavior, which affects the PI lengths for different nominal levels (see supplementary materials for details).
A noticeable difference between the network classes is that the ED models result in sharper PIs than the ST models. 
This coincides with the empirical PI coverages of the methods in that wider PIs typically result in a higher coverage.
Fig.~\ref{fig:eval:calibration} shows the PIT histograms of the postprocessed forecasts. 
While differences are seen between DRN-type and BQN-type models, all DRN-type and all BQN-type models show very similar patterns. While all models are well calibrated, DRN-type models reveal limitations in the resolution of gusts in the lower segment of the distribution. BQN-type models all yield very uniform calibration histograms. 

\begin{table*}[t] 
\caption{Mean CRPS in m/s, PI length in m/s, and PI coverage in \% of the postprocessing methods for the different lead times of the wind gust data (20-member ensemble, year 2016).
Recall that the nominal level of the PIs is approximately 90.48\%. The best-performing models (wrt.\ CRPS) are marked in bold.
\label{tab:eval:gusts}}
\centering
\smallskip
\input{table_results_gusts.txt}
\end{table*} 

\subsection{EUPPBench surface temperature reforecasts}

As shown in Table~\ref{tab:eval:eupp:reforecasts}, both ED and ST models show significant advantages compared to the EPS and EMOS in terms of CRPS and PI length for the EUPPBench dataset. Differences between the network variants arise mainly due to the use of different forecast distribution types.
Note that the lead times of the wind gust dataset are in the short range with a maximum of 18h, whereas the lead times considered in the EUPPBench dataset range from one to five days. Hence, the differences between the lead times in the effects of postprocessing are more pronounced. 
E.g., for a lead time of 120h, the improvement of the network-based postprocessing methods over the conventional EMOS approach is much smaller than for shorter lead times.
In particular, ST models perform the best for lead time 24h and all newly proposed models result in the smallest CRPS for lead time 120h.
In terms of the PI length and coverage, we find that the ED and ST models tend to generate slightly sharper predictions.
A more detailed discussion of the differences in the PI lengths due to the choice of the underlying distribution is provided in the supplementary material.
The PIT histograms in Fig.~\ref{fig:eval:calibration} show that the BQN models struggle to set accurate upper and lower bounds for the predicted distribution, whereas DRN distributions do not show such issues. 
Instead, they face the problem that the tail is too heavy.
Overall, all postprocessing methods result in calibrated forecasts, while the DRN forecasts appear slightly better calibrated than the BQN forecasts, yielding PIT histograms with a wave-like structure.

\begin{table*}[t] 
\caption{Mean CRPS in K, PI length in K, and PI coverage in \% of the postprocessing methods for the different lead times for the EUPPBench reforecast data (11-member ensemble, years 2013-2016). 
Recall that the nominal level of the PIs is approximately 83.33\,\%. The best-performing models (wrt.\ CRPS) are marked in bold. 
\label{tab:eval:eupp:reforecasts}}
\centering
\smallskip
\input{table_results_eupp_reforecasts.txt}
\end{table*}

\subsection{Generalization to 51-member forecast ensembles}

As before, postprocessing outperforms the EPS forecasts and results in calibrated and accurate forecasts (cf.~Table~\ref{tab:eval:eupp:forecasts} and Fig.~\ref{fig:eval:calibration}).
Notably, all models have been trained purely on 11-member reforecasts and are not fine-tuned to the 51-member forecast ensembles.
The CRPS scores are similar with almost identical values for all models, except EMOS, for all lead times. The ST models again perform the best for the shortest lead time.
For the DRN forecasts, we find that the ensemble-based networks tend to reduce the PI length, as it is smaller for all cases except for lead time 120h. The corresponding PI coverages are closely connected to the length of the PIs and indicate that the PIs are too large for most postprocessing models, as the observed coverages are above the nominal level.

The calibration of the methods is not as good as in the other case studies, as indicated by the PIT histograms in Fig.~\ref{fig:eval:calibration}, which may be a consequence of the large learning rate used in training the models (cf.\ supplementary materials). All BQN forecasts have problems in the tails, where the lower and upper bound are too extreme, such that insufficiently many observations fall into the outer bins. DRN yields similar results as for the reforecast data with too heavy-tailed forecast distribution, as indicated by the least frequent last bin. The differences between the methods themselves are again minor. Still, all postprocessing methods generate reasonably well-calibrated forecasts.
Overall, the ensemble-based models result in state-of-the-art performance for generalization on 51-member forecasts or offer advantages over the summary-based benchmark methods.

\begin{table*}[t] 
\caption{Mean CRPS in K, PI length in K, and PI coverage in \% of the postprocessing methods for the different lead times for EUPPBench forecast data (51-member ensemble, years 2017-2018). 
Recall that the nominal level of the PIs is approximately 96.15\%. The best-performing models (wrt.\ CRPS) are marked in bold.
\label{tab:eval:eupp:forecasts}
}
\centering
\smallskip
\input{table_results_eupp_forecasts.txt}
\end{table*}

\section{Analysis of predictor importance\label{sec:features}}

We analyze how the different model types distill relevant information out of the ensemble predictors. For this, we propose an ensemble-oriented permutation feature importance (PFI) analysis to assess which distribution properties of the ensemble-valued predictors have the most effect on the final prediction. 
In its original form, PFI \citep[e.g.,][]{breiman2001random, Rasp2018, Schulz2022} is used to assign relevance scores to scalar-valued predictors by randomly shuffling the values of a single predictor across the dataset. While the idea of shuffling predictor samples translates identically from scalar-valued to ensemble-valued predictors, ensemble predictors possess internal degrees of freedom (DOFs), such as ensemble mean and ensemble range, which may affect the prediction differently. 
In addition to ensemble-internal DOFs, the perturbed predictor ensemble is embedded in the context of the remaining multivariate ensemble predictors, such that covariances, copulas or the rank order of the ensemble members may carry information. To account for such effects, we introduce a conditional permutation strategy that singles out the effects of different ensemble properties. 

\subsection{Importance of the ensemble information}

Following the notation of Section~\ref{sec:benchmark}, let $g: \left[\mathbb{R}^p\right]_M \rightarrow \Theta$ denote a postprocessing system that translates a raw ensemble forecast $X = \left\{\bm{x}_1, ..., \bm{x}_M\right\} \in \left[\mathbb{R}^p\right]_M$ into a postprocessed distribution descriptor $\theta\in \Theta$, and let for each member forecast $\bm{x}_m$ the forecast value of the $i$-th predictor be $\bm{x}_m^{(i)} \in \mathbb{R}$ (for $i = 1, ..., p$). Given a test dataset consisting of known raw forecast-observation pairs, as well as a (negatively oriented) accuracy measure $\bar{S}$, such as the expected CRPS, we write $\bar{S}[g]$ to denote the accuracy score of $g$ on the test data. In this notation, the relative PFI, as used in \cite{Schulz2022}, can be written as 
\begin{equation}
    \Delta_0(P) := \frac{\bar{S}[g \circ P] - \bar{S}[g]}{\bar{S}[g]}\text{,}
    \label{eq:relpi}
\end{equation}
wherein $P$ indicates a perturbation operator that alters parts of the predictor data, and $\circ$ denotes function composition. For the classical PFI, we denote the permutation operator as $\Pi^{(i)}_\pi$, which shuffles the $i$-th predictor channel of the raw ensembles according to a permutation $\pi$ of the dataset.

For ensemble-valued predictors, we consider two generalizations of this operator. We refer to these as the fully-random permutation, $\Pi^{(i)}_\pi$, and the rank-aware random permutation, $\tilde{\Pi}^{(i)}_\pi$. The former acts as a direct analog of the scalar-valued permutation case, i.e., given a dataset $\mathcal{D}:=\left\{(X(t), y(t)): t = 1, ..., T\right\} \subseteq [\mathbb{R}^p]_M \times \mathbb{R}$ of forecast-observation pairs,
it replaces for all $m = 1, ..., M$ the values $\bm{x}_m^{(i)}$ of the ensemble $X\!(t)$ with arbitrary values $\bm{x}_{m'}^{(i)}$, $1 \leq m' \leq M$, from the ensemble $X\!(\pi(t))$, without replacement.
Thus, it destroys all information of the original ensemble. 
The latter ranks the member values $\bm{x}_m^{(i)}$ in $X\!(t)$ and replaces them with values $\bm{x}_{m'}^{(i)}$ from $X\!(\pi(t))$, where $m'$ are chosen such that all members are used exactly once and the perturbed ensemble possesses the same ranking order as the original one. 
It thus preserves the ordering of the perturbed predictors in the context of the remaining predictors. In practice, we note that the differences in feature importance for both variants are very minor, such that we select only the rank-aware variant for further analysis. 

To probe the importance of ensemble-internal DOFs, we consider additional perturbation operators, which rely on conditional shuffling of the ensemble predictors.
For this, let $s: [\mathbb{R}]_M \rightarrow \mathbb{R}$ be a summary function, 
which translates an ensemble of scalar predictor values into a real-valued summary statistic, such as ensemble mean or standard deviation. 
Then an $s$-conditional shuffling operator $\Pi_{\{\pi_b\}|s}^{(i)}$ is defined as follows. For all raw predictions $X\!(t)$ in the dataset, the predictor ensemble for the $i$-th predictor, $X^{(i)}\!(t) = \{\bm{x}_m^{(i)}: \bm{x}_m \in X\!(t)\}$, is extracted and summary statistics $s(X^{(i)}\!(t))$ are computed. The observed summary statistics are ranked from $1$ to $T$ and the corresponding ensembles $X\!(t)$ are distributed into $B\in\mathbb{N}$ evenly spaced bins, according to these ranks. For each bin $b$, $0 \le b < B$, a permutation $\pi_b$ is sampled randomly and the values of the $i$-th predictor are shuffled bin-wise according to these permutations. For suitably sized bins, the shuffling preserves information about $s$ and erases information about other DOFs, thereby ensuring that each of the bins contains an approximately equal number of samples, independent of the details of the predictor distribution. 
Empirically, $B = 100$ bins yielded a good balance between information preservation and randomization. Results for larger and smaller bin sizes were qualitatively similar. Note that for predictors in which certain values appear with large multiplicities, such as zero in censored variables like precipitation, the ranking is computed on the unique values of the summary statistics. This enforces a small amount of variation even in bins with degenerate values.
In analogy to the rank-aware (unconditional) shuffling, the rank-aware $s$-conditional shuffling is denoted as $\tilde{\Pi}_{ \{\pi_b\}|s}^{(i)}$. For the conditional PFI analysis, we suggest the computation of importance ratios,
\begin{equation}
    \chi(P | R) := \frac{\bar{S}[g \circ P] - \bar{S}[g]}{\bar{S}[g \circ R] - \bar{S}[g]}\text{,}
\end{equation}
which measure the fraction of skill restored (or destroyed) by applying a shuffling operation $P$ instead of a reference operation $R$. The ratios of interest are $\chi(\tilde{\Pi}_{\{\pi_b\}|s}^{(i)}, \tilde{\Pi}^{(i)}_\pi)$, which measure how much of the prediction skill deficit due to randomized shuffling of predictor $i$ is restored by preserving information about the summary statistic $s$. In absence of sampling errors due to finite data, $\chi\left(\tilde{\Pi}_{\{\pi_b\}|s}^{(i)}, \tilde{\Pi}^{(i)}_\pi\right)$ yields values between 0 and 1, with 0 indicating uninformative summary statistics, and 1 suggesting that knowledge of $s$ is sufficient to restore the original model skill entirely. 
Empirically, we find that the theoretical bounds are preserved well for predictors with sufficiently large PFI.

\subsection{Results}

\begin{figure*}
    \centering
    \includegraphics[width=\textwidth]{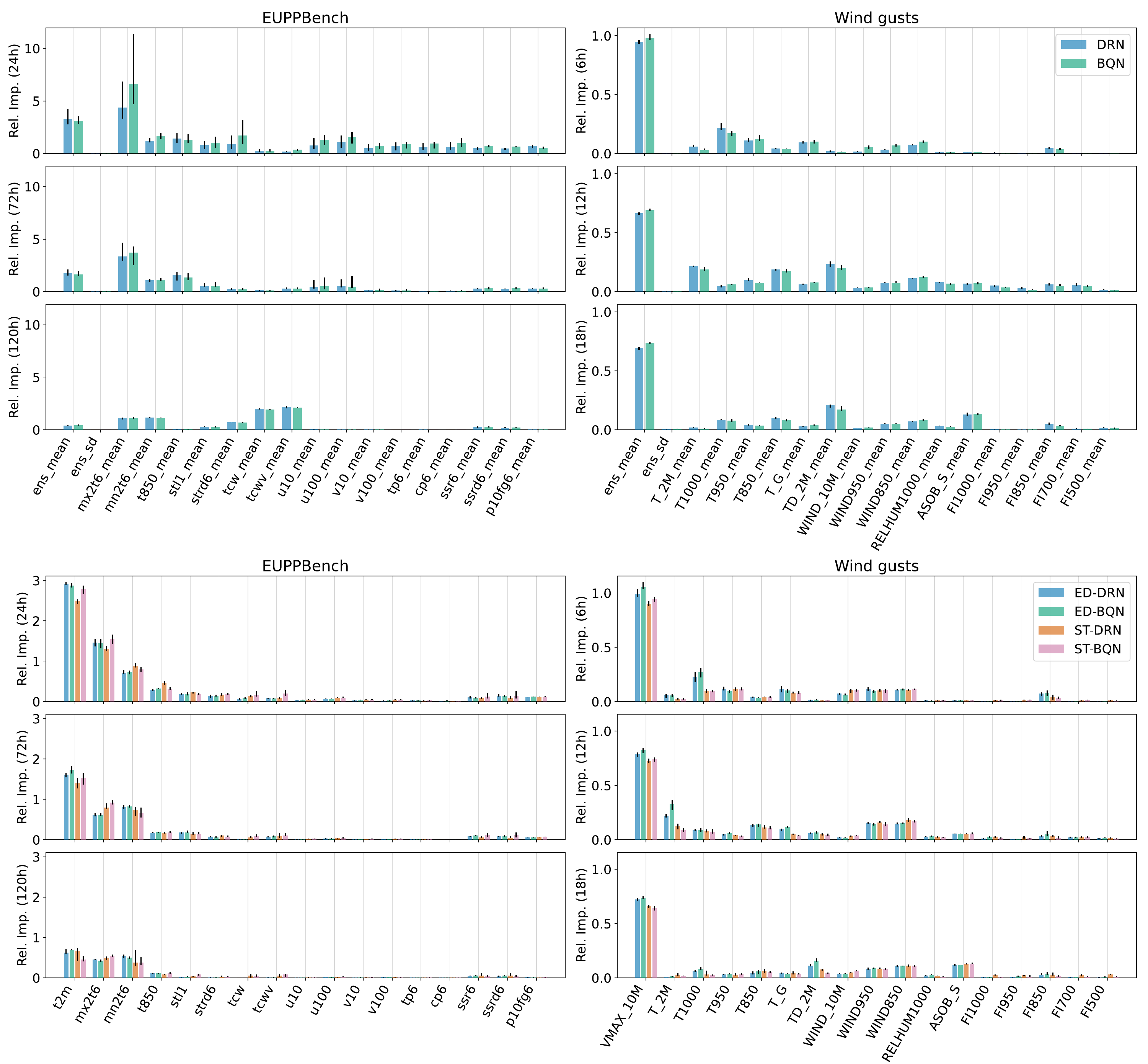}\\
    \caption{Permutation feature importance for summary-based networks (top) and permutation-invariant models (bottom) for EUPPBench and wind gust postprocessing. Bar heights indicate the median of an ensemble of 20 separate models, the error bars depict the IQR. Predictors named \emph{ens} in the top figure correspond to the primary predictors t2m and VMAX-10M, respectively. The suffix \emph{sd} indicates the ensemble standard deviation of the predictor.}
    \label{fig:importance:permutation}
\end{figure*}

We compute PFI scores $\Delta_0\left(\Pi^{(i)}_\pi\right)$ for all ensemble predictors and model variants. Fig.~\ref{fig:importance:permutation} depicts a selection of the PFI scores of the most important ensemble-valued predictors in both tasks. A figure with all ensemble-valued predictors is shown in the supplementary materials.
Scalar-valued predictors (cf.\ Section~\ref{sec:benchmark}\ref{sec:auxpred} for the terminology) are omitted to simplify comparisons with the conditional importance measures. The bar charts show the median of ratios obtained from 20 separate model runs, which have been evaluated independently, and the error bars indicate the interquartile range (IQR). 

The accuracy of the wind gust models is dominated by VMAX-10M, and supplemented by additional predictors with lower importance. Temperature-like predictors obtain similar or higher scores than, e.g.\ winds at 850 hPa and 950 hPa pressure levels. Note that for each lead time, the importance highlights different temperature predictors, which may be attributed to the diurnal cycle. 
Similar arguments can explain the increasing importance of ASOB-S (short-wavelength radiation balance at the surface) with increasing lead time. In a direct comparison of the model variants, we find that the differences between BQN-type and DRN-type models are very minor. However, ED-type models attribute higher importance to the most relevant predictors (VMAX-10M, T1000, T-2M), whereas ST-type models distribute the importance more evenly and use more diverse predictor information. 

In the EUPPBench case, the models focus mainly on temperature-like predictors as well as surface radiation balances. Notably, for the summary-based models, mn2t6 and mx2t6 tend to be more important than the primary predictor t2m up to lead time 72h.  Since the diurnal cycle does not cause variations between the lead times here, differences in the predictor utilization must be due to the increasing uncertainty at longer lead times. The ensemble-based models rely relatively more strongly on the t2m predictor for the shorter lead time, whereas for longer lead times, the information utilization is more diverse. Qualitative differences between ED- and ST-type models are observed with respect to the humidity-related predictors tcw and tcwv. Only ST models recognize the value in these predictors, which may explain in parts the different generalization properties of ED and ST models on the EUPPBench reforecast and forecast datasets. 

\begin{figure*}
    \centering
    \includegraphics[width=\textwidth]{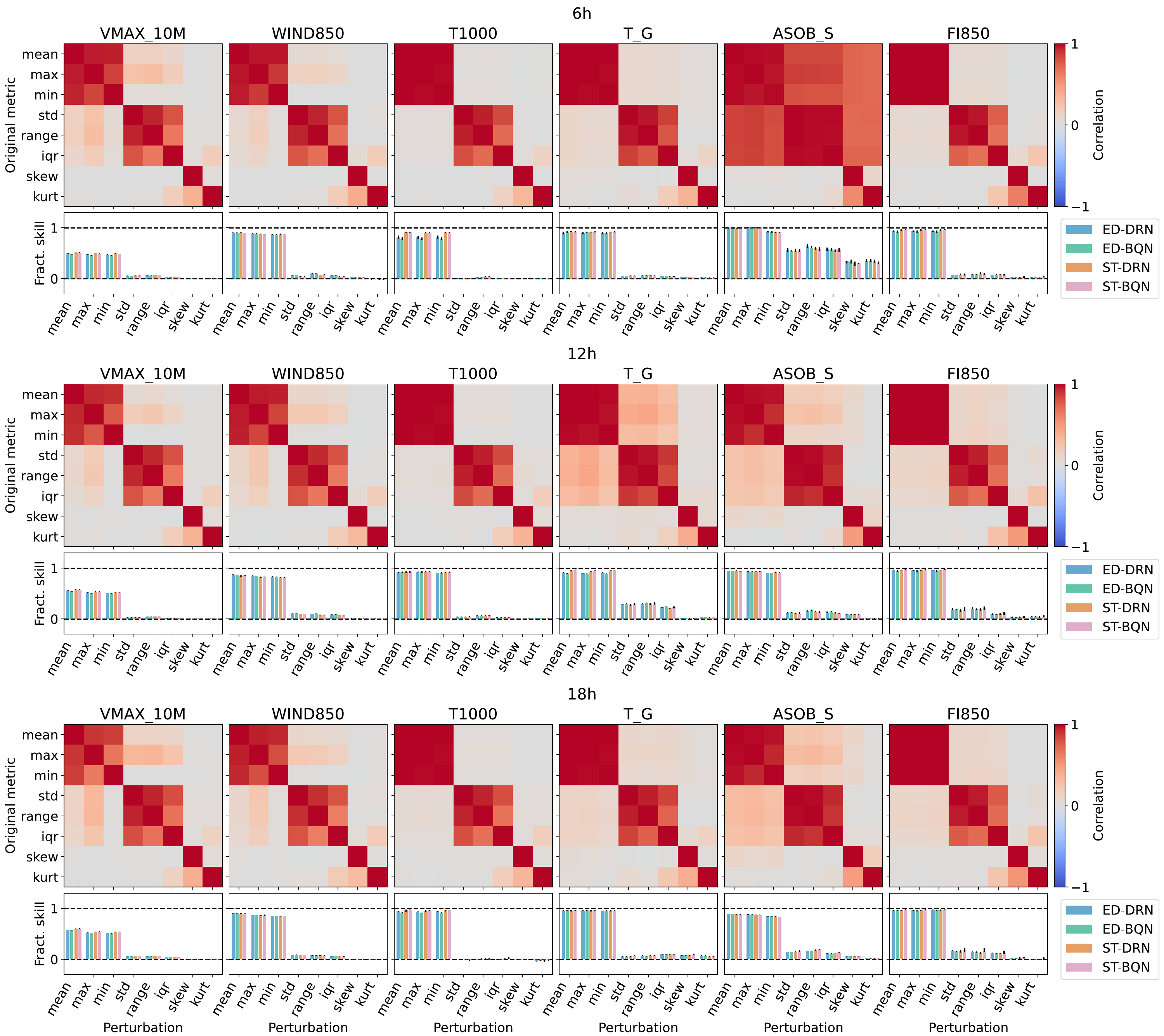}
    \caption{Importance of ensemble-internal DOFs for wind gust postprocessing. Bar charts show importance ratios $\chi(\tilde{\Pi}_{\{\pi_b\}|s}^{(i)}, \tilde{\Pi}^{(i)}_\pi)$ for selected summary statistics $s$, and heatmaps display the Spearman rank correlation between the summary statistics computed on the original dataset and the same statistics after conditional shuffling with respect to the different summary statistics. Bar heights indicate the median of an ensemble of 20 separate models, the error bars depict the IQR.}
    \label{fig:importance:ensdofs:gusts}
\end{figure*}

\begin{figure*}
    \centering
    \includegraphics[width=\textwidth]{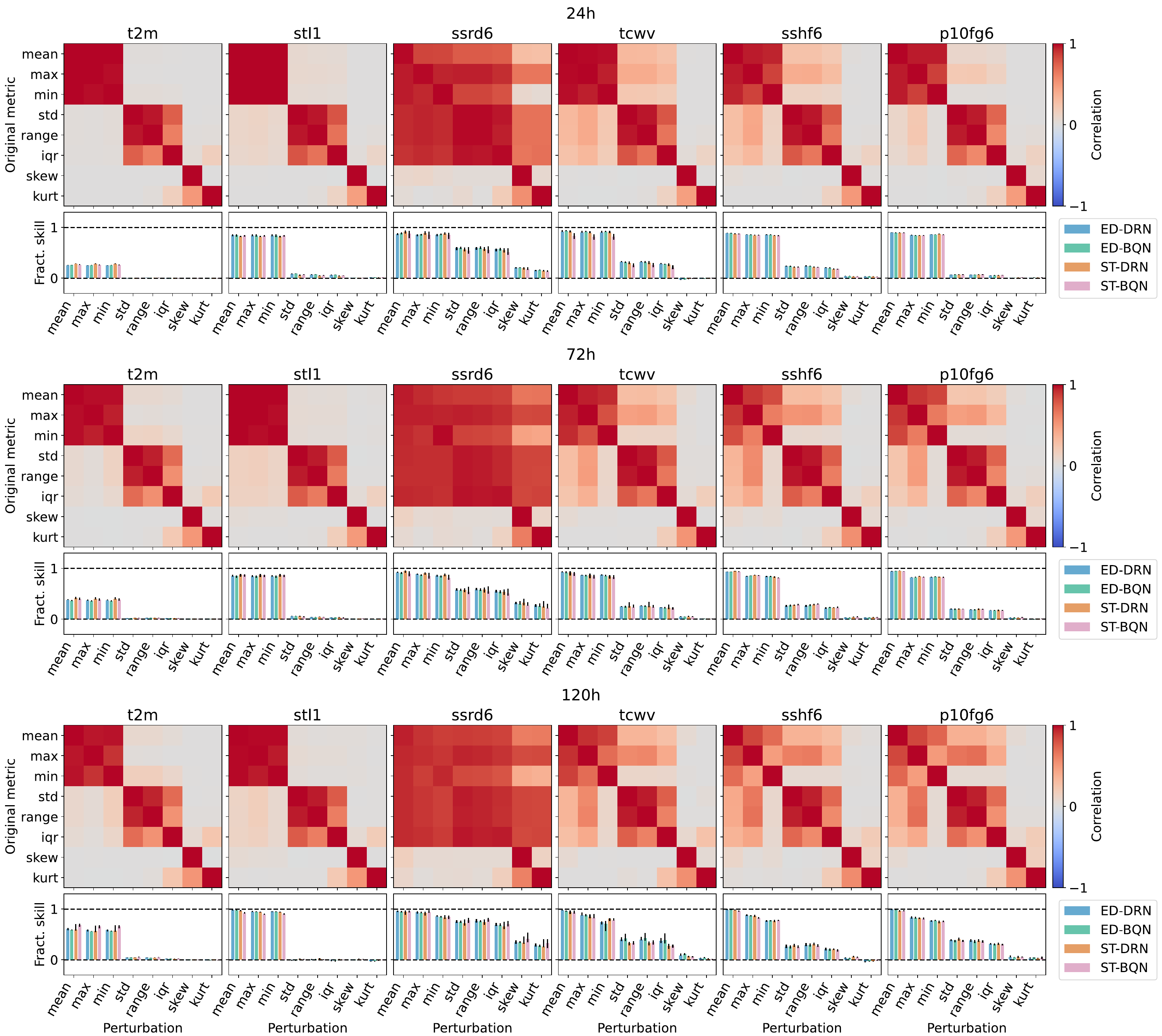}
    \caption{Importance of ensemble-internal DOFs for temperature postprocessing. Same as Fig.~\ref{fig:importance:ensdofs:gusts}. 
    }
    \label{fig:importance:ensdofs:eupp}
\end{figure*}

Figs.~\ref{fig:importance:ensdofs:gusts} and~\ref{fig:importance:ensdofs:eupp} investigate the importance of ensemble-internal DOFs of selected ensemble predictors for the permutation-invariant model architectures. For both datasets, we choose a set of representative high-importance predictors and display the DOF importance for the ensemble-based models. Corresponding figures for the remaining predictors are shown in the supplementary materials. For all predictors and lead times, we compute importance ratios $\chi\left(\tilde{\Pi}_{\{\pi_b\}|s}^{(i)}, \tilde{\Pi}^{(i)}_\pi\right)$ for a selection of commonly used ensemble summary statistics. Specifically, we consider the ensemble mean as a proxy for the location of the distribution, ensemble maximum and minimum to assess the impact of extreme values, standard deviation, IQR, and full range (difference between maximum and minimum) to quantify the scale of the distribution, as well as skewness and kurtosis as higher-order summary statistics.  
Due to the pairwise similarity of some of the measures, it is to be expected that conditional shuffling with respect to one of the measures preserves information about others. To assess the information overlap between shuffling patterns with different reference statistics, Spearman rank correlations are computed between the shuffled statistics and the original statistics. The resulting correlation matrices illustrate how accurately the rank order for one statistic is preserved if the data is conditionally shuffled with respect to another. Rank correlations are chosen to minimize the effect of the marginal distribution of the respective statistics values, since these may vary considerably between different predictors and summary statistics. The results are depicted as heatmaps in Figs.~\ref{fig:importance:ensdofs:gusts} and~\ref{fig:importance:ensdofs:eupp}. 

For wind gust postprocessing (Fig.~\ref{fig:importance:ensdofs:gusts}), the importance ratios suggest in many cases that most of the predictor information can be restored by conditioning the shuffling procedure on the ensemble mean. This is the case for T-1000, T-G and FI850. The interaction plots suggest that the mean-conditioning preserves information about extrema to a high degree, whereas ensemble range and higher-order statistic information are randomized. These findings are supported by observations in \cite{Schulz2022}, who note that omitting the standard deviation of the auxiliary ensemble predictors helps to improve the quality of the network predictions. Larger importance ratios of the scale-related and higher-order DOFs are observed, e.g., for T-G at lead time 12h and ASOB-S at lead time 6h. However, these cases coincide with increased correlations between the respective perturbation patterns and the location-preserving perturbations, which show fractional skill ratios close to unity. This may be seen as an indication that the relevance of the remaining DOFs must in part be attributed to information overlap with the location-related DOFs. Note here that the strong information overlap between location-like and scale-like metrics for ASOB-S predictors at 6h lead time is again an artifact due to the diurnal cycle. At 6h lead time, a substantial fraction of the ASOB-S predictor ensembles fall to zero mean and no variance due to the lack of solar irradiation, which impacts the correlation values as well as the effectiveness of perturbations. 
WIND850 is a corner case, in which the mean-conditioning restores substantial amounts of the model skill, but fails to restore the unperturbed performance completely. This indicates that, while the ensemble mean is an important predictor, the remaining DOFs deliver complementary information that modulates the interpretation of the mean value. 
VMAX-10M, being the primary predictor, constitutes the only example for which no single predictor is sufficient to restore the unperturbed model skill, thus indicating that both ED- and ST-type models learn to attend to the details of the ensemble distribution.

In surface temperature postprocessing, t2m is the primary predictor and displays similar characteristics as VMAX-10M in the wind-gust study.
The mean-conditional shuffling of t2m tends to become more effective in restoring the model skill with increasing lead time. This may be due to the decreasing reliability of the EPS-based predictor ensembles with increasing lead time. Similar patterns are observed also in the remaining predictors. While the model skill cannot be restored with mean-only conditioning for 24h lead time, the mean appears to become more informative for longer lead times. The radiation parameter ssrd6 sticks out visually with high correlations between location-related predictors, which occurs due to the same reasons as for the ASOB-S parameter discussed before. 

%% file: table_results_gusts.txt
\begin{tabular}{c@{\hskip 0.5cm}c@{\hskip 0.3cm}c@{\hskip 0.3cm}c@{\hskip 0.5cm}c@{\hskip 0.3cm}c@{\hskip 0.3cm}c@{\hskip 0.5cm}c@{\hskip 0.3cm}c@{\hskip 0.3cm}c} 
\toprule 
Lead Time & \multicolumn{3}{c}{6h} & \multicolumn{3}{c}{12h} & \multicolumn{3}{c}{18h} \\ 
\midrule 
Method & CRPS & PI length & PI coverage & CRPS & PI length & PI coverage & CRPS & PI length & PI coverage \\ 
\midrule 
EPS & 1.31 & 2.37 & 43.18 & 1.26 & 3.31 & 56.32 & 1.32 & 3.80 & 59.78 \\ 
EMOS & 0.88 & 5.58 & 92.83 & 0.97 & 6.01 & 91.92 & 1.04 & 6.43 & 92.46 \\ 
\midrule 
BQN & $\bm{0.79}$ & 4.60 & 90.23 & $\bm{0.85}$ & 4.90 & 89.65 & $\bm{0.95}$ & 5.56 & 90.70 \\ 
DRN & $\bm{0.79}$ & 4.75 & 91.43 & $\bm{0.85}$ & 5.11 & 91.08 & $\bm{0.95}$ & 5.68 & 91.78 \\ 
\midrule 
ED-BQN & 0.80 & 4.56 & 89.83 & 0.86 & 4.92 & 89.56 & $\bm{0.95}$ & 5.55 & 90.55 \\ 
ED-DRN & $\bm{0.79}$ & 4.70 & 91.17 & 0.86 & 5.15 & 91.13 & $\bm{0.95}$ & 5.76 & 92.07 \\ 
ST-BQN & 0.80 & 4.67 & 90.20 & 0.87 & 5.01 & 89.94 & 0.96 & 5.61 & 90.70 \\ 
ST-DRN & 0.80 & 4.77 & 91.34 & 0.86 & 5.17 & 91.13 & 0.96 & 5.83 & 92.24 \\ 
\bottomrule 
\end{tabular}

%% file: table_results_eupp_reforecasts.txt
\begin{tabular}{c c c c c c c c c c}
     \toprule
    Lead Time & \multicolumn{3}{c}{24h} & \multicolumn{3}{c}{72h} & \multicolumn{3}{c}{120h} \\
    \midrule
    Method & CRPS & PI length & PI coverage & CRPS & PI length & PI coverage & CRPS & PI length & PI coverage \\
    \midrule
    EPS &  1.21 & 1.81 & 39.85 & 1.28 & 3.29 & 56.62 & 1.54 & 4.89 & 63.44 \\
    % EMOS (global) & 1.02 & 4.24 & 79.31 & 1.13 & 4.99 & 80.66 & 1.39 & 6.24 & 80.67 \\
    EMOS & 0.82 & 3.85 & 82.56 & 0.96 & 4.72 & 83.96 & 1.25 & 6.05 & 83.12 \\
    \midrule
    BQN & 0.67 & 3.32 & 84.73 & 0.87 & 4.44 & 86.08 & 1.20 & 6.24 & 86.09 \\
    DRN & 0.67 & 3.28 & 84.16 & $\bm{0.86}$ & 4.27 & 84.58 & $\bm{1.19}$ & 5.70 & 83.09 \\
    \midrule
    ED-BQN & 0.67 & 3.29 & 84.57 & 0.87 & 4.45 & 86.05 & $\bm{1.19}$ & 6.03 & 85.45 \\
    ED-DRN & 0.67 & 3.19 & 83.39 & 0.87 & 4.25 & 84.11 & $\bm{1.19}$ & 5.64 & 82.60 \\
    ST-BQN & $\bm{0.66}$ & 3.16 & 84.01 & 0.87 & 4.31 & 84.85 & $\bm{1.19}$ & 6.07 & 85.15 \\
    ST-DRN & $\bm{0.66}$ & 3.06 & 82.67 & 0.87 & 4.18 & 83.44 & $\bm{1.19}$ & 5.77 & 83.17 \\
    \bottomrule
\end{tabular}

%% file: table_results_eupp_forecasts.txt
\begin{tabular}{c c c c c c c c c c}
     \toprule
    Lead Time & \multicolumn{3}{c}{24h} & \multicolumn{3}{c}{72h} & \multicolumn{3}{c}{120h} \\
    \midrule
    Method & CRPS & PI length & PI coverage & CRPS & PI length & PI coverage & CRPS & PI length & PI coverage \\
    \midrule
    EPS &  1.21 & 2.65 & 57.54 & 1.18 & 4.71 & 74.78 & 1.38 & 7.14 & 83.26 \\
    EMOS & 0.79 & 6.31 & 96.26 & 0.90 & 7.74 & 97.49 & 1.16 & 9.92 & 97.47 \\
    \midrule
    BQN & 0.64 & 4.32 & 94.13 & $\bm{0.80}$ & 6.52 & 97.23 & 1.13 & 9.18 & 97.58 \\
    DRN & 0.64 & 5.48 & 97.92 & $\bm{0.80}$ & 7.21 & 98.37 & 1.13 & 9.58 & 98.28 \\
    \midrule
    ED-BQN & 0.64 & 4.74 & 96.30 & 0.81 & 6.49 & 97.42 & $\bm{1.12}$ & 8.81 & 97.15 \\
    ED-DRN & 0.64 & 5.31 & 97.62 & 0.81 & 7.09 & 98.19 & $\bm{1.12}$ & 9.61 & 97.90 \\
    ST-BQN & $\bm{0.62}$ & 4.61 & 95.96 & $\bm{0.80}$ & 6.18 & 96.31 & 1.13 & 8.68 & 96.05 \\
    ST-DRN & $\bm{0.62}$ & 5.10 & 97.40 &  0.81 & 6.88 & 97.55 & 1.13 & 9.43 & 97.11 \\
    % DRN & 0.64 & 5.42 & 97.59 & $\bm{0.80}$ & 6.88 & 97.73 & 1.17 & 9.70 & 97.57 \\
    \bottomrule
\end{tabular}

%% file: discussion.tex
\section{Discussion and Conclusion\label{sec:conclusion}}

We have introduced permutation-invariant NN architectures for postprocessing ensemble forecasts by selecting two exemplary model families and adapting them to postprocessing. In two case studies, using datasets for wind gusts and surface temperature postprocessing, we evaluated the model performance and compared the permutation-invariant models against benchmark models from prior work. Our results show that permutation-invariant postprocessing networks achieve state-of-the-art performance in both applications. All permutation-invariant architectures outperform both the raw ensemble forecast and conventional postprocessing via EMOS by a large margin, but systematic differences between the (more complex) permutation-invariant models and existing NN-based solutions are very minor and can mostly be attributed to differences in the distribution parameterization. Qualitatively similar results were observed for extreme events in both case studies, but are not shown explicitly in the interest of brevity.

Based on a subsequent assessment of the permutation importance of ensemble-internal DOFs, we have seen that for many auxiliary ensemble predictors, preserving information about the ensemble mean is sufficient to maintain almost the complete information about the postprocessing target, while more detailed information is required about the primary predictors. These findings are consistent with prior work and are more comprehensive due to the larger variety of summary statistics considered in the analysis. 

A striking advantage of the permutation-invariant models lies in the generality of the approach, i.e., the models possess the flexibility of attending to the important features in the predictor ensembles and the capability of identifying those during training (as shown in our feature analysis). As the added flexibility comes with a surplus of computational complexity, the benefits of the respective methods should be weighed carefully. In operational settings, it may be reasonable to consider permutation-invariant models, as proposed here, as a tool for identifying relevant aspects of the input data. The gained knowledge can then be used for data reduction and to train reduced models with a more favorable accuracy-complexity trade-off. 

Despite this, the apparent similarity between the performance of the ensemble-based and summary-based models remains baffling and requires further clarification.
Supposing capable ensemble predictions, it seems reasonable, from a meteorological perspective, to expect that postprocessing models that operate on the entire ensemble can learn more complex patterns and relationships than models that operate on simple summary statistics. The lack of substantial improvements, as seen in this study, admits different explanations. One possibility would be that the available datasets are insufficient to establish statistically relevant connections between higher-order ensemble-internal patterns and the predicted variables. Problems could arise, e.g.\ due to insufficient sample counts of the overall datasets or due to ensemble sizes being too low to provide reliable representations of the forecast distribution. Yet, another reason could lie in the fact that the generation mechanisms underlying the NWP ensemble forecasts fail to achieve meaningful representations of such higher-order distribution information, which would raise follow-up questions regarding the design of future ensemble prediction systems. Given the impact and potential implications of the latter alternative, future work should examine the information content of raw ensemble predictions in more detail. The proposed permutation-invariant model architectures may help to achieve this, e.g., by conducting postprocessing experiments with dynamical toy systems that are cheap to simulate and simple to understand, such that large datasets can be generated and evidence for both hypotheses can be distinguished.

%% file: acknowledgments.tex
This research was funded by the subprojects B5 and C5 of the Transregional Collaborative Research Center SFB/TRR 165 “Waves to Weather” (www.wavestoweather.de) funded by the German Research Foundation (DFG). Sebastian Lerch gratefully acknowledges support by the Vector Stiftung through the Young Investigator Group "Artificial Intelligence for Probabilistic Weather Forecasting". 
We thank two anonymous reviewers for their constructive comments.

%% file: appendix.tex
\appendix[A]
\appendixtitle{Description of predictors}

\label{app:predictors}

The descriptions of the ensemble-valued predictor variables used in both case studies are shown in Tables~\ref{tab:predictors:gusts} and~\ref{tab:predictors:eupp} for wind-gust and surface-temperature postprocessing, respectively. The predictors listed in Table~\ref{tab:predictors:joint} are not ensemble-valued and are used equally in both case studies.

\begin{table*}[h]
    \centering
    \caption{Description of meteorological parameters for wind-gust postprocessing \cite[cf.][]{Schulz2022}. Target variable: wind speed of gust (observations). Primary predictor: VMAX-10m (ensemble forecast).}
    \begin{tabular}{l c l l}
    \toprule
    Short name & Units & Full name & Levels \\
    \midrule
    VMAX & $\text{m}/\text{s}$ & Maximum wind, i.e. wind gusts & 10 m \\
    U & $\text{m}/\text{s}$ & U-component of wind & 10 m, 1000 hPa, 950 hPa, 850 hPa, 700 hPa, 500 hPa \\
    V & $\text{m}/\text{s}$ & V-component of wind & 10 m, 1000 hPa, 950 hPa, 850 hPa, 700 hPa, 500 hPa \\
    WIND & $\text{m}/\text{s}$ & Wind speed, derived from U and V via $\sqrt{\text{U}^2 + \text{V}^2}$ & 10 m, 1000 hPa, 950 hPa, 850 hPa, 700 hPa, 500 hPa\\ 
    OMEGA & $\text{Pa}/\text{s}$ & Vertical velocity (Pressure) & 1000 hPa, 950 hPa, 850 hPa, 700 hPa, 500 hPa\\
    \midrule
    T & K & Temperature & Ground-level, 2 m, \\ 
    & & & 1000 hPa, 950 hPa, 850 hPa, 700 hPa, 500 hPa \\
    T-D & K & Dew point temperature & 2 m \\
    \midrule
    RELHUM & \% & Relative humidity & 1000 hPa, 950 hPa, 850 hPa, 700 hPa, 500 hPa \\ 
    TOT-PREC & $\text{kg}/\text{m}^2$ & Total precipitation (Acc.) & -- \\
    RAIN-GSP & $\text{kg}/\text{m}^2$ & Large scale rain (Acc.) & -- \\
    SNOW-GSP & $\text{kg}/\text{m}^2$ & Large scale snowfall - water equivalent (Acc.) & -- \\
    W-SNOW & $\text{kg}/\text{m}^2$ & Snow depth water equivalent & -- \\
    W-SO & $\text{kg}/\text{m}^2$ & Column integrated soil moisture & multilayers: 1, 2, 6, 18, 54 \\
    CLC & \% & Cloud cover & T: total; \\
    & & & L: soil to 800 hPa; M: 800 to 400 hPa; H: 400 to 0 hPa \\
    HBAS-SC & m & Cloud base above mean sea level, shallow connection & -- \\
    HTOP-SC & m & Cloud top above mean sea level, shallow connection  & -- \\
    \midrule
    ASOB-S & $\text{W} / \text{m}^2$ & Net short wave radiation ﬂux & surface \\
    ATHB-S & $\text{W} / \text{m}^2$  & Net long wave radiation ﬂux (m) & surface \\
    ALB-RAD & \% & Albedo (in short-wave) & -- \\
    \midrule
    PMSL & Pa & Pressure reduced to mean sea level & -- \\
    FI & $\text{m}^2 / \text{s}^2$ & Geopotential & 1000 hPa, 950 hPa, 850 hPa, 700 hPa, 500 hPa\\ 
    \bottomrule
    \end{tabular}
    \label{tab:predictors:gusts}
\end{table*}

\begin{table*}[]
    \centering
    \caption{Description of meteorological parameters for surface temperature postprocessing \citep[EUPPBench, cf.][]{demaeyer2023euppbench}. Target variable: t2m (observations). Primary predictor: t2m (ensemble forecast).}
    \begin{tabular}{l c l l}
        \toprule
        Short name & Units & Full name & Levels \\
        \midrule
        t & K & Temperature & 2\,m, 850 hPa \\
        mx2t6 & K & Max. temperature (6h preceding) & 2\,m\\
        mn2t6 & K & Min. temperature (6h preceding) & 2\,m\\
        \midrule
        z & $\text{m}^2 / \text{s}^2$ & Geopotential & 500 hPa \\
        u & $\text{m}/\text{s}$ & U-component of wind & 10\,m, 100\,m, 700 hPa \\
        v & $\text{m}/\text{s}$ & V-component of wind & 10\,m, 100\,m, 700 hPa \\
        p10fg6 & $\text{m}/\text{s}$ & Max. wind gust in the last 6 hours & 10\,m  \\
        \midrule
        q & $\text{kg} / \text{kg}$ & Specific humidity & 700 hPa \\
        r & \% & Relative humidity & 850 hPa \\
        cape & $\text{J} / \text{kg}$ & Convective available potential energy & -- \\
        cin & $\text{J} / \text{kg}$ & Convective inhibition & -- \\
        tp6 & m & Total precipitation  (6h preceding) & --\\
        cp6 & m & Convective precipitation  (6h preceding) & -- \\
        tcw & $\text{kg} / \text{m}^2$ & Total column water & -- \\
        tcwv & $\text{kg} / \text{m}^2$ & Total column water vapor & -- \\
        tcc & 0 - 1 & Total cloud cover & -- \\
        vis & m & Visibility & -- \\
        \midrule
        sshf6 & $\text{J} / \text{m}^2$ & 	Surface sensible heat flux (6h preceding) & --\\
        slhf6 & $\text{J} / \text{m}^2$ & Surface latent heat flux (6h preceding) & --\\
        ssr6 & $\text{J} / \text{m}^2$ & Surface net short-wave (solar) radiation (6h preceding) & --\\
        ssrd6 & $\text{J} / \text{m}^2$ & Surface net short-wave (solar) radiation downwards (6h preceding) & --\\
        str6 & $\text{J} / \text{m}^2$ & Surface net long-wave (thermal) radiation (6h preceding) & --\\
        strd6 & $\text{J} / \text{m}^2$ & Surface net long-wave (thermal) radiation downwards (6h preceding) & --\\
        \midrule
        swv & $\text{m}^3 / \text{m}^3$ & Volumetric 
        soil water & l1: 0 - 7 cm \\
        sd & m & Snow depth - water equivalent & -- \\
        st & K & Soil temperature & l1: 0 - 7 cm\\
        \midrule
        \bottomrule
    \end{tabular}
    \label{tab:predictors:eupp}
\end{table*}

\begin{table*}[]
    \centering
    \caption{Auxiliary predictors for both datasets \citep[cf.][]{Schulz2022}.}
    \begin{tabular}{l c l}
        \toprule
        Predictor & Type & Description \\
        \midrule
        yday & Temporal & Cosine transformed day of the year \\
        lat & Spatial & Latitude of the station \\
        lon & Spatial & Longitude of the station \\
        alt & Spatial & Altitude of the station \\
        orog & Spatial & Diﬀerence of station altitude and model surface height of nearest grid point. \\
        loc-bias & Spatial & Mean bias of ensemble forecasts, computed from the training data. \\
        loc-cover & Spatial & Mean coverage of ensemble forecasts, computed from the training data. \\
        \bottomrule
    \end{tabular}
    \label{tab:predictors:joint}
\end{table*}

% \clearpage

\appendix[B]

\appendixtitle{Model hyperparameters}

Table~\ref{tab:tuning:final} displays the hyperparameter settings for all model configurations used in the experiments. For details about the hyperparameter tuning process, we refer to the supplementary materials.

\begin{table*}[h]
    \renewcommand{\arraystretch}{1.}
    \centering
    \caption{Hyper parameters for model experiments.}
    \input{table_hyperparams.txt}
    \label{tab:tuning:final}
\end{table*}

\clearpage

%% file: table_hyperparams.txt
\begin{tabular}{l l c c c c c c}
    \toprule
    &  & \multicolumn{6}{c}{Settings}  \\
    & & \multicolumn{3}{c}{Wind gusts} & \multicolumn{3}{c}{EUPPBench} \\
    Method & Parameter & 6h & 12h & 18h & 24h & 72h & 120 h \\ \midrule
    Common & optimizer & \multicolumn{6}{c}{Adam} \\
    & batch size & \multicolumn{6}{c}{64 for DRN and BQN, 128 else} \\
    & hidden layers & \multicolumn{6}{c}{2 hidden layers per MLP, 3 transformer blocks} \\
    & dimension of station embedding & \multicolumn{6}{c}{10} \\
    & activation (at output layer) & \multicolumn{6}{c}{softplus (softplus)} \\
    & DRN / EMOS distribution family & \multicolumn{6}{c}{truncated logistic} \\
    & training epochs & \multicolumn{3}{c}{150} & \multicolumn{3}{c}{250} \\ \midrule
    DRN & channels (first layer / second layer) & \multicolumn{3}{c}{64 / 32} & \multicolumn{3}{c}{64 / 32} \\
    & learning rate & \multicolumn{3}{c}{$5\times10^{-4}$} & $3\times10^{-3}$ & $3\times10^{-3}$ & $4\times10^{-4}$ \\
    & patience & \multicolumn{3}{c}{10} & \multicolumn{3}{c}{24} \\ \midrule
    BQN & channels (first layer / second layer) & \multicolumn{3}{c}{48 / 24} & \multicolumn{3}{c}{48 / 24} \\
    & polynomial degree & \multicolumn{3}{c}{12} & \multicolumn{3}{c}{12} \\
    & learning rate & \multicolumn{3}{c}{$5\times10^{-4}$} & $4\times10^{-3}$ & $4\times10^{-3}$ & $3\times10^{-4}$ \\
    & patience & \multicolumn{3}{c}{10} & \multicolumn{3}{c}{24} \\ \midrule
    ED-DRN & channels (encoder) & \multicolumn{3}{c}{64} & \multicolumn{3}{c}{64} \\
    & channels (decoder) & \multicolumn{3}{c}{64} & \multicolumn{3}{c}{48} \\
    & dropout (encoder)    & \multicolumn{3}{c}{0.02} & \multicolumn{3}{c}{0.05} \\
    & dropout (decoder)    & \multicolumn{3}{c}{0.00} &  0.05 &  0.05 &  0.10 \\
    & learning rate        & \multicolumn{3}{c}{$4\times10^{-4}$} &  $2\times10^{-4}$ & $2\times10^{-4}$ & $1\times10^{-4}$ \\
    & patience            & \multicolumn{3}{c}{24} &  20 &  20 & 15 \\ \midrule
    ED-BQN & channels (encoder) &  \multicolumn{3}{c}{64} & \multicolumn{3}{c}{64} \\
    & channels (decoder) & \multicolumn{3}{c}{64} & \multicolumn{3}{c}{48} \\
    & dropout (encoder)  & \multicolumn{3}{c}{0.00} & 0.05 & 0.10 &   0.10 \\
    & dropout (decoder)  & \multicolumn{3}{c}{0.00} & 0.05 & 0.10 & 0.10 \\
    & polynomial degree & \multicolumn{3}{c}{12} & \multicolumn{3}{c}{8} \\
    & learning rate      & \multicolumn{3}{c}{$2\times10^{-4}$} &  $5\times10^{-4}$ & $2\times10^{-4}$ & $2\times10^{-4}$ \\
    & patience           & \multicolumn{3}{c}{24} & 20 & 20 & 12 \\ \midrule
    ST-DRN & channels (transformer) &  \multicolumn{3}{c}{64} & \multicolumn{3}{c}{64} \\
    & channels (decoder) & \multicolumn{3}{c}{64} & \multicolumn{3}{c}{48} \\
    & dropout (transformer) &  \multicolumn{3}{c}{0.02} &  \multicolumn{3}{c}{0.00} \\
    & dropout (decoder) &  \multicolumn{3}{c}{0.05} &  \multicolumn{3}{c}{0.00} \\
    & learning rate     & $2\times10^{-4}$ & $1\times10^{-4}$ & $5\times10^{-4}$ & $2\times10^{-4}$ & $1\times10^{-4}$ & $5\times10^{-4}$ \\
    & patience & \multicolumn{3}{c}{24} & \multicolumn{3}{c}{24} \\ \midrule
    ST-BQN & channels (transformer) & \multicolumn{3}{c}{64} & \multicolumn{3}{c}{48} \\
    & channels (decoder) & \multicolumn{3}{c}{64} & \multicolumn{3}{c}{48} \\
    & dropout (transformer) & \multicolumn{3}{c}{0.10} & 0.20 & 0.25 & 0.20 \\
    & dropout (decoder)    & \multicolumn{3}{c}{0.00} &  0.05 & 0.10 & 0.15 \\
    & polynomial degree & \multicolumn{3}{c}{12} & \multicolumn{3}{c}{8} \\
    & learning rate        & \multicolumn{3}{c}{$1\times10^{-4}$} & \multicolumn{3}{c}{$5\times10^{-5}$} \\
    & patience            & \multicolumn{3}{c}{24} & 24 & 15 & 20 \\
    \bottomrule
\end{tabular}

%% file: main.bbl
\begin{thebibliography}{57}
\providecommand{\natexlab}[1]{#1}
\providecommand{\url}[1]{\texttt{#1}}
\renewcommand{\UrlFont}{\rmfamily}
\providecommand{\urlprefix}{URL }
\expandafter\ifx\csname urlstyle\endcsname\relax
  \providecommand{\doi}[1]{https://doi.org/\discretionary{}{}{}#1}\else
  \providecommand{\doi}{https://doi.org/\discretionary{}{}{}\begingroup
  \urlstyle{rm}\Url}\fi
\providecommand{\eprint}[2][]{\url{#2}}

\bibitem[{Baldauf et~al.(2011)Baldauf, Seifert, F{\"{o}}rstner, Majewski,
  Raschendorfer,, and Reinhardt}]{Baldauf2011}
Baldauf, M., A.~Seifert, J.~F{\"{o}}rstner, D.~Majewski, M.~Raschendorfer, and
  T.~Reinhardt, 2011: {Operational convective-scale numerical weather
  prediction with the COSMO model: Description and sensitivities}.
  \textit{Monthly Weather Review}, \textbf{139~(12)}, 3887--3905,
  \doi{10.1175/MWR-D-10-05013.1}.

\bibitem[{Ben-Bouallegue et~al.(2023)Ben-Bouallegue, Weyn, Clare, Dramsch,
  Dueben,, and Chantry}]{ben2023improving}
Ben-Bouallegue, Z., J.~A. Weyn, M.~C. Clare, J.~Dramsch, P.~Dueben, and
  M.~Chantry, 2023: Improving medium-range ensemble weather forecasts with
  hierarchical ensemble transformers. \textit{arXiv preprint arXiv:2303.17195}.

\bibitem[{Breiman(2001)}]{breiman2001random}
Breiman, L., 2001: Random forests. \textit{Machine learning}, \textbf{45},
  5--32.

\bibitem[{Bremnes(2020)}]{Bremnes2020}
Bremnes, J.~B., 2020: {Ensemble postprocessing using quantile function
  regression based on neural networks and Bernstein polynomials}.
  \textit{Monthly Weather Review}, \textbf{148~(1)}, 403--414,
  \doi{10.1175/mwr-d-19-0227.1}.

\bibitem[{Burkart and Huber(2021)Burkart, and Huber}]{burkart2021survey}
Burkart, N., and M.~F. Huber, 2021: A survey on the explainability of
  supervised machine learning. \textit{Journal of Artificial Intelligence
  Research}, \textbf{70}, 245--317, \doi{10.1613/jair.1.12228}.

\bibitem[{Chen et~al.(2022)Chen, Janke, Steinke,, and Lerch}]{Chen2022}
Chen, J., T.~Janke, F.~Steinke, and S.~Lerch, 2022: Generative machine learning
  methods for multivariate ensemble post-processing. \textit{arXiv preprint
  arXiv:2211.01345}.

\bibitem[{Dai and Hemri(2021)Dai, and Hemri}]{Hemri2021}
Dai, Y., and S.~Hemri, 2021: Spatially coherent postprocessing of cloud cover
  ensemble forecasts. \textit{Monthly Weather Review}, \textbf{149~(12)}, 3923
  -- 3937, \doi{10.1175/MWR-D-21-0046.1}.

\bibitem[{Demaeyer et~al.(2023)}]{demaeyer2023euppbench}
Demaeyer, J., and Coauthors, 2023: The euppbench postprocessing benchmark
  dataset v1.0. \textit{Earth System Science Data Discussions}, \textbf{2023},
  1--25, \doi{10.5194/essd-2022-465}.

\bibitem[{ECMWF(2013)}]{climetlab2022}
ECMWF, 2013: {C}li{M}et{L}ab. GitHub, \url{https://github.com/ecmwf/climetlab}.

\bibitem[{Edwards and Storkey(2016)Edwards, and Storkey}]{edwards2016towards}
Edwards, H., and A.~Storkey, 2016: Towards a neural statistician. \textit{arXiv
  preprint arXiv:1606.02185}.

\bibitem[{Farokhmanesh et~al.(2023)Farokhmanesh, H{\"o}hlein,, and
  Westermann}]{farokhmanesh2023deep}
Farokhmanesh, F., K.~H{\"o}hlein, and R.~Westermann, 2023: Deep learning--based
  parameter transfer in meteorological data. \textit{Artificial Intelligence
  for the Earth Systems}, \textbf{2~(1)}, e220\,024,
  \doi{10.1175/AIES-D-22-0024.1}.

\bibitem[{Finn(2021)}]{finn2021self}
Finn, T.~S., 2021: Self-attentive ensemble transformer: Representing ensemble
  interactions in neural networks for earth system models. \textit{arXiv
  preprint arXiv:2106.13924}.

\bibitem[{Gneiting et~al.(2007)Gneiting, Balabdaoui,, and
  Raftery}]{Gneiting2007probabilistic}
Gneiting, T., F.~Balabdaoui, and A.~E. Raftery, 2007: {Probabilistic forecasts,
  calibration and sharpness}. \textit{Journal of the Royal Statistical Society.
  Series B: Statistical Methodology}, \textbf{69~(2)}, 243--268,
  \doi{10.1111/j.1467-9868.2007.00587.x}.

\bibitem[{Gneiting and Katzfuss(2014)Gneiting, and Katzfuss}]{Gneiting2014}
Gneiting, T., and M.~Katzfuss, 2014: {Probabilistic forecasting}.
  \textit{Annual Review of Statistics and Its Application}, \textbf{1~(1)},
  125--151, \doi{10.1146/annurev-statistics-062713-085831}.

\bibitem[{Gneiting and Raftery(2007)Gneiting, and
  Raftery}]{Gneiting2007scoring}
Gneiting, T., and A.~E. Raftery, 2007: {Strictly proper scoring rules,
  prediction, and estimation}. \textit{Journal of the American Statistical
  Association}, \textbf{102~(477)}, 359--378, \doi{10.1198/016214506000001437}.

\bibitem[{Gneiting et~al.(2005)Gneiting, Raftery, Westveld,, and
  Goldman}]{Gneiting2005}
Gneiting, T., A.~E. Raftery, A.~H. Westveld, and T.~Goldman, 2005: {Calibrated
  probabilistic forecasting using ensemble model output statistics and minimum
  CRPS estimation}. \textit{Monthly Weather Review}, \textbf{133~(5)},
  1098--1118, \doi{10.1175/MWR2904.1}.

\bibitem[{Gneiting and Ranjan(2011)Gneiting, and Ranjan}]{Gneiting2011}
Gneiting, T., and R.~Ranjan, 2011: Comparing density forecasts using threshold-
  and quantile-weighted scoring rules. \textit{Journal of Business \& Economic
  Statistics}, \textbf{29~(3)}, 411--422, \doi{10.1198/jbes.2010.08110}.

\bibitem[{Gr{\"o}nquist et~al.(2021)Gr{\"o}nquist, Yao, Ben-Nun, Dryden,
  Dueben, Li,, and Hoefler}]{gronquist2021deep}
Gr{\"o}nquist, P., C.~Yao, T.~Ben-Nun, N.~Dryden, P.~Dueben, S.~Li, and
  T.~Hoefler, 2021: Deep learning for post-processing ensemble weather
  forecasts. \textit{Philosophical Transactions of the Royal Society A},
  \textbf{379~(2194)}, 20200\,092, \doi{10.1098/rsta.2020.0092}.

\bibitem[{Guidotti et~al.(2018)Guidotti, Monreale, Ruggieri, Turini,
  Giannotti,, and Pedreschi}]{guidotti2018survey}
Guidotti, R., A.~Monreale, S.~Ruggieri, F.~Turini, F.~Giannotti, and
  D.~Pedreschi, 2018: A survey of methods for explaining black box models.
  \textit{ACM computing surveys (CSUR)}, \textbf{51~(5)}, 1--42.

\bibitem[{Haupt et~al.(2021)Haupt, Chapman, Adams, Kirkwood, Hosking, Robinson,
  Lerch,, and Subramanian}]{Haupt2021}
Haupt, S.~E., W.~Chapman, S.~V. Adams, C.~Kirkwood, J.~S. Hosking, N.~H.
  Robinson, S.~Lerch, and A.~C. Subramanian, 2021: Towards implementing
  artificial intelligence post-processing in weather and climate: proposed
  actions from the oxford 2019 workshop. \textit{Philosophical Transactions of
  the Royal Society A: Mathematical, Physical and Engineering Sciences},
  \textbf{379~(2194)}, 20200\,091, \doi{10.1098/rsta.2020.0091}.

\bibitem[{H{\"o}hlein et~al.(2020)H{\"o}hlein, Kern, Hewson,, and
  Westermann}]{hohlein2020comparative}
H{\"o}hlein, K., M.~Kern, T.~Hewson, and R.~Westermann, 2020: A comparative
  study of convolutional neural network models for wind field downscaling.
  \textit{Meteorological Applications}, \textbf{27~(6)}, e1961,
  \doi{10.1002/met.1961}.

\bibitem[{Horat and Lerch(2023)Horat, and Lerch}]{horat_lerch_2023}
Horat, N., and S.~Lerch, 2023: Deep learning for post-processing global
  probabilistic forecasts on sub-seasonal time scales. \textit{arXiv preprint
  arXiv:2306.15956}.

\bibitem[{Höhlein(2023)}]{hoehlein2023code}
Höhlein, K., 2023: Code {"Postprocessing of Ensemble Weather Forecasts Using
  Permutation-invariant Neural Networks"}. Zenodo,
  \doi{10.5281/zenodo.8329345}.

\bibitem[{Jordan et~al.(2019)Jordan, Kr{\"{u}}ger,, and Lerch}]{Jordan2019}
Jordan, A., F.~Kr{\"{u}}ger, and S.~Lerch, 2019: {Evaluating probabilistic
  forecasts with scoringRules}. \textit{Journal of Statistical Software},
  \textbf{90~(12)}, 1--37, \doi{10.18637/jss.v090.i12}.

\bibitem[{Khan et~al.(2022)Khan, Naseer, Hayat, Zamir, Khan,, and
  Shah}]{khan2022transformers}
Khan, S., M.~Naseer, M.~Hayat, S.~W. Zamir, F.~S. Khan, and M.~Shah, 2022:
  Transformers in vision: A survey. \textit{ACM computing surveys (CSUR)},
  \textbf{54~(10s)}, 1--41, \doi{10.1145/3505244}.

\bibitem[{Koenker and Bassett(1978)Koenker, and Bassett}]{Koenker1978}
Koenker, R., and G.~Bassett, 1978: Regression quantiles. \textit{Econometrica},
  \textbf{46~(1)}, 33--50, \doi{10.2307/1913643}.

\bibitem[{Labe and Barnes(2021)Labe, and Barnes}]{labe2021detecting}
Labe, Z.~M., and E.~A. Barnes, 2021: Detecting climate signals using
  explainable {AI} with single-forcing large ensembles. \textit{Journal of
  Advances in Modeling Earth Systems}, \textbf{13~(6)}, e2021MS002\,464,
  \doi{10.1029/2021MS002464}.

\bibitem[{Lee et~al.(2019)Lee, Lee, Kim, Kosiorek, Choi,, and Teh}]{Lee2019}
Lee, J., Y.~Lee, J.~Kim, A.~Kosiorek, S.~Choi, and Y.~W. Teh, 2019: Set
  transformer: A framework for attention-based permutation-invariant neural
  networks. \textit{International conference on machine learning}, PMLR,
  3744--3753.

\bibitem[{Lerch and Thorarinsdottir(2013)Lerch, and
  Thorarinsdottir}]{lerch_thorarinsdottir_2013}
Lerch, S., and T.~L. Thorarinsdottir, 2013: Comparison of non-homogeneous
  regression models for probabilistic wind speed forecasting. \textit{Tellus
  A}, \textbf{65~(1)}, 21\,206, \doi{10.3402/tellusa.v65i0.21206}.

\bibitem[{Linardatos et~al.(2020)Linardatos, Papastefanopoulos,, and
  Kotsiantis}]{linardatos2020explainable}
Linardatos, P., V.~Papastefanopoulos, and S.~Kotsiantis, 2020: Explainable
  {AI}: A review of machine learning interpretability methods.
  \textit{Entropy}, \textbf{23~(1)}, 18, \doi{10.3390/e23010018}.

\bibitem[{Lyle et~al.(2020)Lyle, van~der Wilk, Kwiatkowska, Gal,, and
  Bloem-Reddy}]{Lyle2020}
Lyle, C., M.~van~der Wilk, M.~Kwiatkowska, Y.~Gal, and B.~Bloem-Reddy, 2020: On
  the benefits of invariance in neural networks. \textit{arXiv preprint
  arXiv:2005.00178}.

\bibitem[{Matheson and Winkler(1976)Matheson, and Winkler}]{Matheson1976}
Matheson, J.~E., and R.~L. Winkler, 1976: {Scoring rules for continuous
  probability distributions}. \textit{Management Science}, \textbf{22~(10)},
  1087--1096, \doi{10.1287/mnsc.22.10.1087}.

\bibitem[{Messner et~al.(2017)Messner, Mayr,, and Zeileis}]{Messner2017}
Messner, J.~W., G.~J. Mayr, and A.~Zeileis, 2017: {Nonhomogeneous boosting for
  predictor selection in ensemble postprocessing}. \textit{Monthly Weather
  Review}, \textbf{145~(1)}, 137--147, \doi{10.1175/MWR-D-16-0088.1}.

\bibitem[{Mlakar et~al.(2023)Mlakar, Mer{\v{s}}e,, and
  Pucer}]{mlakar2023ensemble}
Mlakar, P., J.~Mer{\v{s}}e, and J.~F. Pucer, 2023: Ensemble weather forecast
  post-processing with a flexible probabilistic neural network approach.
  \textit{arXiv preprint arXiv:2303.17610}.

\bibitem[{Molnar et~al.(2023)Molnar, K{\"o}nig, Bischl,, and
  Casalicchio}]{molnar2023model}
Molnar, C., G.~K{\"o}nig, B.~Bischl, and G.~Casalicchio, 2023: Model-agnostic
  feature importance and effects with dependent features: a conditional
  subgroup approach. \textit{Data Mining and Knowledge Discovery}, 1--39,
  \doi{10.1007/s10618-022-00901-9}.

\bibitem[{Pantillon et~al.(2018)Pantillon, Lerch, Knippertz,, and
  Corsmeier}]{Pantillon2018}
Pantillon, F., S.~Lerch, P.~Knippertz, and U.~Corsmeier, 2018: {Forecasting
  wind gusts in winter storms using a calibrated convection-permitting
  ensemble}. \textit{Quarterly Journal of the Royal Meteorological Society},
  \textbf{144~(715)}, 1864--1881, \doi{10.1002/qj.3380}.

\bibitem[{Rasp and Lerch(2018)Rasp, and Lerch}]{Rasp2018}
Rasp, S., and S.~Lerch, 2018: {Neural networks for postprocessing ensemble
  weather forecasts}. \textit{Monthly Weather Review}, \textbf{146~(11)},
  3885--3900, \doi{10.1175/MWR-D-18-0187.1}.

\bibitem[{Ravanbakhsh et~al.(2016)Ravanbakhsh, Schneider,, and
  Poczos}]{Ravanbakhsh2016}
Ravanbakhsh, S., J.~Schneider, and B.~Poczos, 2016: Deep learning with sets and
  point clouds. \textit{arXiv preprint arXiv:1611.04500}.

\bibitem[{Reichstein et~al.(2019)Reichstein, Camps-Valls, Stevens, Jung,
  Denzler, Carvalhais,, and Prabhat}]{reichstein2019deep}
Reichstein, M., G.~Camps-Valls, B.~Stevens, M.~Jung, J.~Denzler, N.~Carvalhais,
  and f.~Prabhat, 2019: Deep learning and process understanding for data-driven
  earth system science. \textit{Nature}, \textbf{566~(7743)}, 195--204,
  \doi{10.1038/s41586-019-0912-1}.

\bibitem[{Sannai et~al.(2019)Sannai, Takai,, and Cordonnier}]{Sannai2019}
Sannai, A., Y.~Takai, and M.~Cordonnier, 2019: Universal approximations of
  permutation invariant/equivariant functions by deep neural networks.
  \textit{arXiv preprint arXiv:1903.01939}.

\bibitem[{Scheuerer(2014)}]{Scheuerer2014}
Scheuerer, M., 2014: {Probabilistic quantitative precipitation forecasting
  using ensemble model output statistics}. \textit{Quarterly Journal of the
  Royal Meteorological Society}, \textbf{140~(680)}, 1086--1096,
  \doi{10.1002/qj.2183}.

\bibitem[{Scheuerer et~al.(2020)Scheuerer, Switanek, Worsnop,, and
  Hamill}]{Scheuerer2020}
Scheuerer, M., M.~B. Switanek, R.~P. Worsnop, and T.~M. Hamill, 2020: {Using
  artificial neural networks for generating probabilistic subseasonal
  precipitation forecasts over California}. \textit{Monthly Weather Review},
  \textbf{148~(8)}, 3489--3506, \doi{10.1175/MWR-D-20-0096.1}.

\bibitem[{Schulz et~al.(2021)Schulz, Ayari, Lerch,, and Baran}]{Schulz2021}
Schulz, B., M.~E. Ayari, S.~Lerch, and S.~Baran, 2021: {Post-processing
  numerical weather prediction ensembles for probabilistic solar irradiance
  forecasting}. \textit{Solar Energy}, \textbf{220}, 1016--1031,
  \doi{10.1016/j.solener.2021.03.023}.

\bibitem[{Schulz and Lerch(2022{\natexlab{a}})Schulz, and
  Lerch}]{Schulz2022preprint}
Schulz, B., and S.~Lerch, 2022{\natexlab{a}}: Aggregating distribution
  forecasts from deep ensembles. \textit{arXiv preprint arXiv:2204.02291}.

\bibitem[{Schulz and Lerch(2022{\natexlab{b}})Schulz, and Lerch}]{Schulz2022}
Schulz, B., and S.~Lerch, 2022{\natexlab{b}}: {Machine learning methods for
  postprocessing ensemble forecasts of wind gusts: A systematic comparison}.
  \textit{Monthly Weather Review}, \textbf{150~(1)}, 235--257,
  \doi{10.1175/mwr-d-21-0150.1}.

\bibitem[{Soelch et~al.(2019)Soelch, Akhundov, van~der Smagt,, and
  Bayer}]{soelch2019aggregations}
Soelch, M., A.~Akhundov, P.~van~der Smagt, and J.~Bayer, 2019: On deep set
  learning and the choice of aggregations. \textit{Artificial Neural Networks
  and Machine Learning -- ICANN 2019: Theoretical Neural Computation}, Springer
  International Publishing, 444--457, \doi{10.1007/978-3-030-30487-4_35}.

\bibitem[{Strobl et~al.(2008)Strobl, Boulesteix, Kneib, Augustin,, and
  Zeileis}]{strobl2008conditional}
Strobl, C., A.-L. Boulesteix, T.~Kneib, T.~Augustin, and A.~Zeileis, 2008:
  Conditional variable importance for random forests. \textit{BMC
  bioinformatics}, \textbf{9}, 1--11, \doi{10.1007/s10618-022-00901-9}.

\bibitem[{Taillardat et~al.(2016)Taillardat, Mestre, Zamo,, and
  Naveau}]{Taillardat2016}
Taillardat, M., O.~Mestre, M.~Zamo, and P.~Naveau, 2016: {Calibrated ensemble
  forecasts using quantile regression forests and ensemble model output
  statistics}. \textit{Monthly Weather Review}, \textbf{144~(6)}, 2375--2393,
  \doi{10.1175/MWR-D-15-0260.1}.

\bibitem[{van Schaeybroeck and Vannitsem(2015)van Schaeybroeck, and
  Vannitsem}]{VanSchaeybroeck2015}
van Schaeybroeck, B., and S.~Vannitsem, 2015: {Ensemble post-processing using
  member-by-member approaches: Theoretical aspects}. \textit{Quarterly Journal
  of the Royal Meteorological Society}, \textbf{141~(688)}, 807--818,
  \doi{10.1002/qj.2397}.

\bibitem[{Vannitsem et~al.(2018)Vannitsem, Wilks,, and Messner}]{Vannitsem2018}
Vannitsem, S., D.~S. Wilks, and J.~W. Messner, 2018: \textit{{Statistical
  Postprocessing of Ensemble Forecasts}}. Elsevier,
  \doi{10.1016/c2016-0-03244-8}.

\bibitem[{Vannitsem et~al.(2021)}]{Vannitsem2021}
Vannitsem, S., and Coauthors, 2021: {Statistical postprocessing for weather
  forecasts: Review, challenges, and avenues in a big data world}.
  \textit{Bulletin of the American Meteorological Society}, \textbf{102~(3)},
  E681 -- E699, \doi{10.1175/BAMS-D-19-0308.1}.

\bibitem[{Vaswani et~al.(2017)Vaswani, Shazeer, Parmar, Uszkoreit, Jones,
  Gomez, Kaiser,, and Polosukhin}]{Vaswani2017}
Vaswani, A., N.~Shazeer, N.~Parmar, J.~Uszkoreit, L.~Jones, A.~N. Gomez, L.~u.
  Kaiser, and I.~Polosukhin, 2017: Attention is all you need. \textit{Advances
  in Neural Information Processing Systems}, Curran Associates, Inc., Vol.~30.

\bibitem[{Veldkamp et~al.(2021)Veldkamp, Whan, Dirksen,, and
  Schmeits}]{Veldkamp2021}
Veldkamp, S., K.~Whan, S.~Dirksen, and M.~Schmeits, 2021: {Statistical
  postprocessing of wind speed forecasts using convolutional neural networks}.
  \textit{Monthly Weather Review}, \textbf{149~(4)}, 1141--1152,
  \doi{10.1175/MWR-D-20-0219.1}.

\bibitem[{Vinyals et~al.(2015)Vinyals, Bengio,, and Kudlur}]{vinyals2015order}
Vinyals, O., S.~Bengio, and M.~Kudlur, 2015: Order matters: Sequence to
  sequence for sets. \textit{arXiv preprint arXiv:1511.06391}.

\bibitem[{Vogel et~al.(2018)Vogel, Knippertz, Fink, Schlueter,, and
  Gneiting}]{vogel2018skill}
Vogel, P., P.~Knippertz, A.~H. Fink, A.~Schlueter, and T.~Gneiting, 2018: Skill
  of global raw and postprocessed ensemble predictions of rainfall over
  northern tropical africa. \textit{Weather and Forecasting}, \textbf{33~(2)},
  369--388, \doi{10.1175/WAF-D-17-0127.1}.

\bibitem[{Zaheer et~al.(2017)Zaheer, Kottur, Ravanbakhsh, Poczos,
  Salakhutdinov,, and Smola}]{Zaheer2017}
Zaheer, M., S.~Kottur, S.~Ravanbakhsh, B.~Poczos, R.~R. Salakhutdinov, and
  A.~J. Smola, 2017: Deep sets. \textit{Advances in Neural Information
  Processing Systems}, Curran Associates, Inc., Vol.~30.

\bibitem[{Zhang et~al.(2019)Zhang, Hare,, and
  Pr{\"u}gel-Bennett}]{zhang2019fspool}
Zhang, Y., J.~Hare, and A.~Pr{\"u}gel-Bennett, 2019: Fspool: Learning set
  representations with featurewise sort pooling. \textit{arXiv preprint
  arXiv:1906.02795}.

\end{thebibliography}
